\documentclass{article} 
\usepackage{preprint,times}

\usepackage{hyperref}
\usepackage{url}

\usepackage{microtype}
\usepackage{graphicx}
\usepackage{subcaption}
\usepackage{booktabs} 
\usepackage{hyperref}

\usepackage{tcolorbox}
\usepackage{enumitem}

\usepackage{listings} 
\usepackage{xcolor} 
\usepackage{multirow} 
\usepackage{bm}
\usepackage{longtable} 

\usepackage{amsmath}
\usepackage{amssymb}
\usepackage{mathtools}
\usepackage{amsthm}

\usepackage{threeparttable}
\usepackage{float}

\usepackage[capitalize,noabbrev]{cleveref}

\theoremstyle{plain}
\newtheorem{theorem}{Theorem}[section]
\newtheorem{proposition}[theorem]{Proposition}

\theoremstyle{definition}
\newtheorem{definition}[theorem]{Definition}

\theoremstyle{remark}

\newcommand{\myendofproof}[0]{\hfill $\blacksquare$ \newline}
\newcommand{\indep}{\perp \!\!\! \perp}

\usepackage[textsize=tiny]{todonotes}

\title{Why Tabular Foundation Models Excel on \\ Relational Data They Have Never Seen}

\author{Linjie Xu, Yiqun Hu, Jixiang He, \& David Wipf\thanks{Correspondence to linjiexu@hkushxlab.com and davidwipf@gmail.com.}  \\
University of Hong Kong, Shanghai X-Lab}

\begin{document}

\maketitle

\input{def.set}

\vspace*{-0.3cm}
\begin{abstract}
Given a relational database (RDB) storing heterogeneous tabular information, how can we predict missing (or future) values in some target column of interest?  As candidate targets are numerous in enterprise settings, it is preferable to avoid learning a new model from scratch each time there is a new prediction task.  Frozen foundation models (FMs) based on in-context learning (ICL) and RDB-specific pre-training provide a viable solution, but the ideal design still remains largely unresolved.  Perhaps unexpectedly, preliminary evidence has recently shown that existing \textit{single-table FMs} (aka TFMs), when combined with suitable parameter-free relational encoders, can be quite competitive in practice despite \textit{zero exposure to RDB pre-training or relational data of any kind}.  In this work we first empirically confirm this phenomenon with a more diverse array of experiments.  We then turn to our main contribution, namely, surfacing multiple complementary explanations for why TFMs behave so effectively.  These explanations ultimately hinge on innate synergies whereby parameter-free encoders map relational effects to fixed-length embeddings that loosely mimic distributions for which TFMs were originally designed;  in so doing, desirable single-table properties transfer to relational settings (e.g., convergent posterior predictive distributions).  Collectively, our analysis forges new pathways for improving RDB FMs, particularly when we prioritize interpretability and inevitable integration within agentic systems.

\end{abstract}

\vspace*{-0.3cm}
\section{Introduction}
\vspace*{-0.2cm}

Relational databases (RDBs), housing collections of interrelated tables, are indispensable across  enterprise applications and e-commerce platforms \citep{garcia2009database}.  From a machine learning standpoint, the information stored within just a single RDB often contains countless possibilities for predictive modeling, covering targets such as customer retention, user churn, or click-through rates \citep{acquire-valued-shoppers-challenge,motl2015ctu,amazon-reviews,retailrocket}.  Of course retraining a new model from scratch to address each new predictive task of interest may be prohibitively laborious or expensive, which motivates a new class of foundation models (FMs) specifically tailored for RDBs.  Among others, these vary from LLM-based approaches that digest serialized RDB content \citep{wydmuch2024tackling}, to specialized Transformers pre-trained based on in-context learning (ICL) \citep{fey2025kumorfm,hudovernik2026kumorfm,ranjanlarge,wang2026relational}, the common theme in each case being a \textit{frozen} architecture capable of making predictions involving arbitrary unseen RDBs.

As a robust and scalable alternative, it has recently been demonstrated that  \textit{single-table FMs} (aka TFMs), when combined with suitable parameter-free subgraph encoders for capturing multi-table relational context, can perform well without requiring any RDB-specific pre-training \citep{grinsztajn2026tabpfn3,hayler2026advancing,JUICE_paper_2026}.  After presenting background RDB context in Section \ref{sec:background}, our work herein centers around confirming and explaining this unexpected performance via:  

\begin{itemize}[leftmargin=0.5cm]
    
    \item \textbf{Comprehensive demonstration.} We provide thorough empirical comparison of wide-varying RDB foundation models across diverse benchmarking tasks.  These results from Section \ref{sec:experiments} solidify the surprising effectiveness of TFMs repurposed for relational data.

    \item \textbf{Explanatory theory.} In Sections \ref{sec:continuum}--\ref{sec:agents}, we elucidate multiple complementary factors that help explain this phenomenon and broader merits of TFMs.  Specifically, we show that TFM-based approaches: (i) naturally handle varying degrees of relational context, forming a continuum from single-tables to rich RDB schemas; (ii) transfer desirable asymptotic properties to RDB use cases, helping to compensate for mismatches between pre-training prior and practical deployment distributions; (iii) are not notably limited by expressivity concerns, even with parameter-free encoders and no prior RDB exposure, and (iv) integrate particularly well with agentic systems.

\end{itemize}

\vspace*{-0.3cm}
\section{Basics of RDB Foundation Models} \label{sec:background}
\vspace*{-0.2cm}

The goal of an RDB foundation model is to retain applicability across unseen RDBs, ideally with no retraining for each new predictive task.  Beyond direct LLM-prompting approaches \citep{wydmuch2024tackling}, we now discuss recent pipelines designed to natively handle RDB/tabular data.

\vspace*{-0.1cm}
\subsection{Predictive Modeling with Relational Context}
\vspace*{-0.2cm}

Suppose we have training data $\{\bx_{i},y_i \}_{i=1}^{n} = \{\bX,\by \}$ composed of instance feature rows $\bx_{i} \in \calX^{d}$ along with associated instance labels $y_i \in \calY$ for all $i$.  The objective is to learn a parameterized model $f_\theta$ that achieves $f_\theta(\bx_{\tiny \mbox{test}}) \approx y_{\tiny \mbox{test}}$ at any test point $\{\bx_{\tiny \mbox{test}}, y_{\tiny \mbox{test}} \}$, where in practice $y_{\tiny \mbox{test}}$ is unknown.  RDB predictive modeling generalizes this canonical ML scenario by integrating an additional set of auxiliary data tables $\calT =\{\bT^k \}_{k=1}^K$, where  $\bT^k \in \mathfrak{T}^{n_k \times d_k}$ denotes the $k$-th table associated with a given entity type.  Each table row corresponds with a single instance of that entity (e.g., a purchased item), while columns reflect heterogeneous instance attributes (e.g., product information).  Note that w.l.o.g., we also assume $\bT^1 = \bX$.  Lastly, an RDB also includes a set of relations $\calR$ expressed as primary key (PK), foreign key (FK) pairs \citep{garcia2009database}; these may be absorbed into the constituent tables of $\calT$ as additional columns when convenient.  Both supervised methods and FMs leverage these auxiliary tables for making predictions.

\vspace*{-0.2cm}
\subsection{Computing RDB Embeddings}
\vspace*{-0.15cm}
To make use of relational context for predictive tasks, the basic strategy is to first convert a given labeled RDB $\calD := \{\calT,\calR,\by \}$ to a heterogeneous graph $\calG$.  This typically involves treating each table $\bT^k$ as a node type and each row $i$ within a given $\bT^k$ as a node with features $\bt^k_{i}$ \citep{cvitkovic2020supervised,dwivedi2025relational,fey2023relational,zhang2023gfs,zhang2023rdbench}.  Directed edges can then be formed using each PK-FK pair within $\calR$.  We next sample $H$-hop subgraphs or ego-networks $\calG_H(\bx_{i})$ from $\calG$ that are centered at each target row $\bx_{i}$ within $\bX$.  Importantly, sampling for \textit{temporal} RDBs should exclude nodes with time-stamps later than $\bx_{i}$.  These subgraphs form the input to an RDB encoder $g_\phi$, instantiated as some form of graph neural network  or Transformer architecture with (optional) parameters $\phi$.  This facilitates computation of \textit{fixed-length} embeddings $\bz_{i} = g_\phi[ \calG_H(\bx_{i}) ] \in \mathbb{R}^{d_z}$.

\vspace*{-0.1cm}
\subsection{Combining with RDB Prediction Heads} \label{sec:prediction_heads}
\vspace*{-0.05cm}

\vspace*{-0.2cm}

\paragraph{Schema-agnostic models.}  Provided the encoder $g_\phi$ is designed to digest a broad spectrum of input RDB schema using a shared representational form, then it is possible to train a single composite model $q_\theta\big(y ~|~\bz_{\tiny \mbox{test}}) \equiv q_\theta\big(y ~|~ g_\phi\big[ \calG_H(\bx_{\tiny \mbox{test}}) \big]  \big)$ to assign high probability to labels $y = y_{\tiny \mbox{test}}$ (or $y \approx y_{\tiny \mbox{test}}$ for regression tasks) across a range of real-world RDBs \citep{ranjan2026relational,wanggriffin,wu-etal-2025-large}.  Here $\bx_{\tiny \mbox{test}}$ can be viewed as an additional unlabeled row(s) appended to $\bX$. Such models can in principle be applied to new unseen RDBs without further training.

\vspace*{-0.1cm}
\paragraph{Pre-trained ICL-based models.}  A limitation of the schema-agnostic models from above is that the effective context available for making predictions is merely the subgraph $\calG_H(\bx_{\tiny \mbox{test}})$, which in isolation may not contain sufficient information pertaining to broader labeling patterns across a vast RDB.  In-context learning (ICL) provides a natural workaround by granting the prediction head diverse samples of multiple analogous subgraphs extracted from each new RDB of interest.  Specifically, ICL-based architectures derive predictions from the revised distribution
\begin{equation}
  q_\theta\big(y ~|~\bz_{\tiny \mbox{test}}, \calS_n),~\mbox{with}~ \calS_n = \{\bz_{i}, y_i \}_{i=1}^n, ~\bz = g_\phi[ \calG_H(\bx) ], \nonumber
\end{equation}
where $\bz_{\tiny \mbox{test}}$ and each $\bz_{i}$ are obtained by passing unlabeled and labeled rows of $\bX$ through the shared encoder $g_\phi$.  It is also possible to apply $g_\phi$ to revised tabular embeddings $\bU^k \leftarrow \bT^k$ rather than original tables \citep{hudovernik2026kumorfm}.  The prediction head $q_\theta$ is implemented via various Transformer architectures that grant flexibility over $n$ and $d_z$.  Finally, model pre-training is conducted using large volumes of synthetically-generated data, possibly combined with  real-world RDBs \citep{chen2026openrfm,fey2025kumorfm,hudovernik2026kumorfm,kothapalli2026plurel,ranjan2026rt,ranjanlarge}.

\vspace*{-0.1cm}

\paragraph{Retrofitting TFMs with a relational encoder.} Given the growing range of single-table FMs \citep{hollmann2025accurate,jingangtabicl,qu2026tabiclv2,zhang2025mitra,zhang2025limix}, it is natural to consider combining with a \textit{parameter-free} encoder $g_{\tiny \mbox{pf}}$, for flattening RDB neighborhoods into fixed-length tabular features, such that the entire pipeline requires no RDB-specific training/pre-training \citep{grinsztajn2026tabpfn3,hayler2026advancing,JUICE_paper_2026} (a conversion akin to $\bU^k \leftarrow \bT^k$, as adopted by the pre-trained ICL-based model from \citet{hudovernik2026kumorfm} described above, may be optionally considered as an additional pre-processing step, but we do not pursue this possibility further).  Herein we aim to confirm and explain the effectiveness of these simple approaches relative to more complex alternatives.

\begin{figure*}[t]
\centering
\includegraphics[width=1\linewidth]{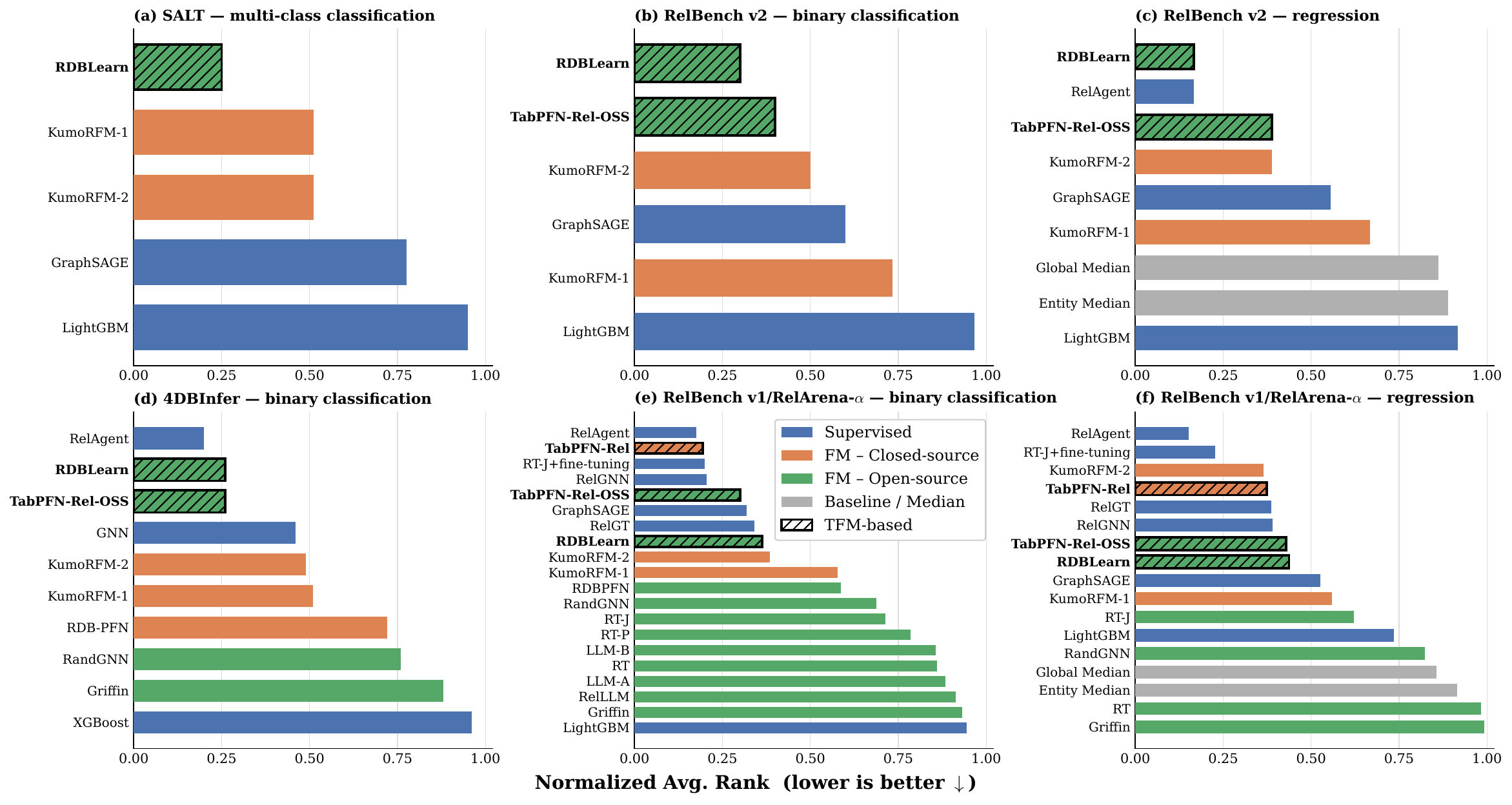}
\vspace*{-0.6cm}
    \caption{A TFM-based approach outperforms other RDB FMs on each benchmark.}
    \label{fig:big_comparison}
  \vspace*{-0.3cm}
\end{figure*}

\vspace*{-0.0cm}

\section{More Empirical Evidence that TFMs Excel on RDBs} \label{sec:experiments}

\vspace*{-0.2cm}

\paragraph{Models.}  For TFM-based approaches, we adopt
\textit{RDBLearn} \citep{JUICE_paper_2026} and \textit{TabPFN-Rel} \citep{hayler2026advancing} based on analogous principles; TabPFN-Rel has both open- and closed-source versions (the latter including proprietary text handling and unknown RDB-related modifications).  For RDB-native FMs, we include a variety of both open-source and closed-source options with published results based on author-specific tuning.  Note that most models only provide results on a subset of available benchmarks.  Lastly, we cover a handful of supervised baselines that represent either SOTA performance or classical methodology as a reference point.  Appendix \ref{app:experiment_details} includes all baseline model descriptions as well as further details of how results were obtained.

\vspace*{-0.2cm}
\paragraph{Results.}  We compare across a suite of six diverse benchmarks from Appendix \ref{app:experiment_details}, each composed of multiple RDB predictive tasks.  We remark that it is routine to only report results on RelBench-v1 datasets/tasks \citep{jager2026tabpfn35,hayler2026advancing,kothapalli2026plurel,ranjan2026relational,ranjanlarge}, which are also equivalent to recent RelArena-$\alpha$ task designations at the time of this writing \citep{hayler2026advancing}.  But this poses greater risk of overfitting to limited benchmarking data, e.g., settings/architectures suitable for RelBench-v1 need not always translate to stronger performance elsewhere.  For the expanded benchmark set considered here, results are shown in Figure \ref{fig:big_comparison} and Appendix \ref{app:extended_results}, where TFM-based approaches achieve the best performance among all FMs, including even several variants that may involve pre-training with RDBs structured similarly to the benchmarks. We stress that RDBLearn and TabPFN-Rel-OSS were run directly from open-source repositories for all benchmarks with no task-specific pipeline adjustments and \textit{zero exposure to RDB data of any kind at any stage}, i.e., no chance for any implicit test-set alignment on public benchmarks.

\vspace*{-0.2cm}
 \paragraph{Why are TFMs so competitive on RDBs?} It is initially reasonable to assume that pre-training RDB-specific FMs using actual RDB data (real or synthetic) is essential for achieving strong inference-time performance on unseen relational data.  So how can TFMs paired with some $g_{\tiny \mbox{pf}}$ possibly perform well, as we have confirmed in Figure \ref{fig:big_comparison}. The high-level argument from \citet{JUICE_paper_2026} is that prior to seeing ICL samples, a data-agnostic encoder is incapable of adjudicating column roles, which may vary from dataset to dataset (or even task to task within a given RDB).  This phenomenon advantages RDB encoders that only compress vertically \textit{within} high-dimensional columns (where shared units facilitate interpretable aggregation), not horizontally \textit{across} columns, where relevance is largely indeterminate without adequate label information.  Conditioned on this restriction to vertical (intra-column) compression, \citet{JUICE_paper_2026}  detail how $g_{\tiny \mbox{pf}}$ can result in minimal loss of expressiveness, while naturally pairing with existing TFMs to form a complete RDB pipeline.  While promising as an analytical starting point, there remains considerable ambiguity regarding the rationale and surprising empirical efficacy of this simple recipe.  In the remainder of this work we take several disambiguating steps, beginning with a discussion of the blurry line between single tables and RDBs, where TFMs already work well on the former.

\vspace*{-0.1cm}
\section{Single Tables and RDBs on a Continuum} \label{sec:continuum}
\vspace*{-0.1cm}

\paragraph{Inducing RDBs from single tables.}  Conventionally, TFMs and benchmarking (i.e., single-table regimes) are treated separately from their RDB counterparts.  But as we now examine, infinitesimal adjustments to an initial single-table setup facilitate gradual transitioning to multi-table relational scenarios, making any strict partitioning somewhat arbitrary.  Concretely, beginning with a single-table scenario, meaning $K=1$ and features $\bT^1 = \bX$, we can trivially transform these data into two or more tables with arbitrarily-small, labeled cross-row relational dependencies as follows:
\vspace*{-0.1cm}
\begin{enumerate}[leftmargin=0.7cm]
    \item Select any categorical column of $\bT^1$; w.l.o.g.~assume $B$ categories.  If no categorical columns exist, then we simply chose a numerical column and quantize into $B$ bins.
    \item Form a new auxiliary table $\bT^2$ with a single length-$B$ column, where elements are the unique categories obtained from step 1 above.  Treat the lone column within $\bT^2$ as a PK, and the original categorical column within $\bT^1$ as an FK.
    \item To induce more complex schema, we may optionally repeat steps 1--2 using other $\bT^1$ columns.

\item Form subgraphs $\calG_H(\bx)$ seeded at any row $\bx$; provided $H \geq 2$, these neighborhoods will include multiple nodes associated with different rows of $\bX$.  This is because the first hop reaches an induced $\bT^k$ ($k>1$), while any second hop returns to $\bT^1$.  In doing so, cross-row relational dependencies are introduced via the edges within $\calG_H(\bx)$ \citep{gan2024graph}. 

\item  Define the label generation function
\vspace*{-0.1cm}
\begin{equation} \label{eq:label_function}
   \hspace*{0.0cm}  y = \psi\big[ (1-\tau) f_1(\bx, \varepsilon_1) + \tau f_2\big( \calG_H(\bx) \backslash \bx,\varepsilon_2 \big) \big],
    \vspace*{-0.1cm}
\end{equation}
where $\tau \in [0,1]$ represents a continuation parameter and $\{\varepsilon_1,\varepsilon_2\} \in \mathbb{R}^2$ are exogenous noise factors.  Meanwhile, $f_1$ and $f_2$ are functions mapping their respective domains to a shared $r$-dimensional embedding space $\mathbb{R}^r$, after which $\psi : \mathbb{R}^r \rightarrow \calY$ converts to the final label space.

\end{enumerate}

\vspace*{-0.2cm}
\paragraph{Continuum perspective.} Any possible conditional distribution $p\left(y | \calG_H(\bx) \right)$ can in principle be obtained by generating data via step 5, subject to mild regularity conditions \citep{Peters2017}[Proposition 4.1]; likewise for special cases $p(y|\bx)$ (when $\tau = 0$) and $p\left(y | \calG_H(\bx) \backslash \bx \right)$ (when $\tau = 1$).  Hence given any set of tabular features, we can always generate a continuum of rich label distributions with varying degree of relational dependency, ranging from the canonical single-table regime ($\tau = 0$) to an interconnected multi-table structure ($\tau \neq 0$). We emphasize that our point here is not to replicate the most likely real-world processes by which RDBs are generated, nor discount the countless other ways single tables can be normalized into RDB-like schema. Instead, we are merely demonstrating that there is \textit{no unambiguous line between single tables and multi-table relational data}, as relational effects can diversely form a continuum from non-existence to prominence.  

\vspace*{-0.2cm}
\paragraph{Plausibility of TFMs on RDBs.} TFMs by original design excel handling $\tau = 0$ cases, and some real-world RDBs likely approximate this regime \citep{akhter2026fair}.  Furthermore, if they are equipped with a suitably powerful encoder $g_{\tiny \mbox{pf}}$, such as the encoding strategy endorsed by \citet{JUICE_paper_2026}, then larger $\tau$ cases are also within reach as evidenced by Figure \ref{fig:big_comparison}.  Hence, TFMs are reasonable candidates for navigating relational continuums, providing a strong initial plausibility argument for their usage on RDBs.  Conversely, if an alternative RDB FM does not work well on single-tables, what about performance on RDBs that are minimally different from single tables, and so on?  Again, these considerations motivate RDB FMs that explicitly handle a broad relational spectrum as TFMs already do. In Appendix \ref{app:ensembling_limitation} we closely examine limitations of adopting interpolation/ensembling between single-table FMs and RDB FMs as a simplistic workaround, proving a pessimistic bound (Proposition \ref{prop:interpolation_limitation}) that occurs even with very minor modeling constraints.  See also Appendix \ref{app:benchmarking_implications} for broader benchmarking implications of the continuum perspective.

\begin{figure*}[t]
\centering
\includegraphics[width=1.0\linewidth]{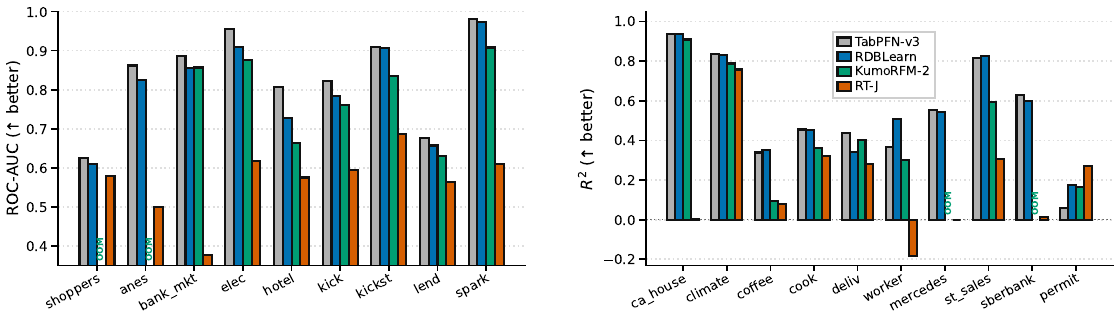}
\vspace*{-0.6cm}
    \caption{Classification/Regression performance on RDBs induced from single-table temporal data.}
    \label{fig:single_table}
  \vspace*{-0.2cm}
\end{figure*}

\vspace*{-0.2cm}
\paragraph{Empirical testing.}  We empirically probe implications of the continuum perspective, particularly the behavior of RDB FMs operating near single-table regimes.  For this purpose, we first collect a set of tabular datasets (originally single table), whereby train/test splits are temporal as in typical RDB scenarios; all setup details are deferred to Appendix \ref{app:continuum_experiment}.  We then induce RDBs using steps 1--4 from above, but we retain the original dataset labels to maintain real-world authenticity; see Appendix \ref{app:semi_syn_experiments} for complementary experiments whereby original labels are synthetically varied across different $\tau$ values within (\ref{eq:label_function}).  Note that these datasets, designed to test non-IID temporal shifts, may already exhibit small degrees of cross-row dependency even without artificially inducing such dependencies via (\ref{eq:label_function}).  We compare TFM-based RDBLearn against the top RDB FMs (non-TFM-based) KumoRFM-2 \citep{hudovernik2026kumorfm} and RT-J \citep{ranjan2026rt,ranjanlarge} in Figure \ref{fig:single_table} and Appendix \ref{app:continuum_experiment}.  We also include TabPFN-v3 \citep{grinsztajn2026tabpfn3} as a non-relational baseline (also shared as the RDBLearn base TFM for direct comparison).

Overall, RDBLearn shows the best performance among RDB FMs, demonstrating the ability to mostly discard distracting relational effects that are not necessary for making predictions.  Meanwhile, although RT-J struggles with some of these tasks, KumoRFM-2 performs significantly better (excluding four OOM cases).  This is perhaps not surprising given that KumoRFM-2 pre-training explicitly transitions from single tables to relational data over an undisclosed multi-stage process \citep{hudovernik2026kumorfm}.  Moreover, inspection of the KumoRFM-2 implementation reveals heavy reliance on the TabICL-v2 TFM architecture \citep{qu2026tabiclv2}.  Pure TFM-based methods like RDBLearn and TabPFN-Rel simply take these observations one step further by dispensing with potentially-complex RDB-specific pre-training altogether.  Either way, it is our view that Figure \ref{fig:single_table}-style benchmarking represents an informative hurdle for RDB FMs to pass.  It is also worth noting that on three regression tasks, RDB FMs actually surpass the performance of the non-relational baseline, and the RDBLearn average regression performance marginally exceeds TabPFN-v3 (see Appendix \ref{app:continuum_experiment}). This plausibly indicates the presence of modest relational effects along the continuum on at least some of these data.  These effects are explicitly amplified in the additional experiments from Appendix \ref{app:semi_syn_experiments}.

\vspace*{-0.1cm}
\section{Locality Considerations Favor TFM Usage} \label{sec:locality}
\vspace*{-0.1cm}

We now turn to an RDB generative process that facilitates the study of locality-based necessary conditions for removing bias.  We then find that TFMs, with uniformly drawn ICL samples produced by a suitable $g_{\tiny \mbox{pf}}$, allow us to approximately satisfy these conditions with RDB data.

\vspace*{-0.1cm}
\subsection{Necessary Conditions for Unbiased RDB FMs} \label{sec:unbiased}
\vspace*{-0.1cm}

\begin{definition} \label{def:reverse_time} (\textit{continual RDB generation})  Let $\calP$ denote a set of RDB distribution sequences $\rho := \{ p_n \}_{n=m}^\infty$, while $\calD_n \sim p_n$ is an RDB associated with $\rho$ for any $n > m \geq 0$.  We stipulate that all such $\calD_n$ include a table $\bT^1$ composed of labeled rows $\{ \bx_i, y_i\}_{i=0}^n$.  These labels are assumed to follow the conditional $p\big(y_i | \calG_H(\bx_i) \big)$ that does not vary across $n$, but may vary with different $\rho \in \calP$ (in contrast, we do not prevent the marginal $p\big(\calG_H(\bx_i) \big)$ from varying across $n$ and $\rho$). We treat $\{\calG_H(\bx_0),y_0\} \equiv \{\calG_H(\bx_{\tiny \mbox{test}}),y_{\tiny \mbox{test}}\}$ as a test sample that remains fixed across $n$.

\end{definition}
\vspace*{0.1cm}

By design, Definition \ref{def:reverse_time} allows us to consider distributions of RDBs whereby the overall size and number of ICL samples is growing indefinitely (these may extend into the past if timestamps are involved), but with a fixed evaluation point $\{\calG_H(\bx_{\tiny \mbox{test}}),y_{\tiny \mbox{test}}\}$ (akin to typical IID tabular cases).  And even if real-world RDBs are not generated in exactly this way, with modest assumptions, the same RDB distribution can often be defined/generated in either a forward or backward fashion.  Hence we preserve flexibility with this construction while facilitating asymptotic analyses as follows.

Consider a scenario where, as $n$ becomes large, a given RDB FM achieves two desirable conditions: (i) the variance of the predicted label vanishes, and (ii) the bias relative to the true label distribution, conditioned on an encoder output, tends to zero.  Mathematically, these conditions manifest as
\begin{eqnarray} \label{eq:ideal_asymptotic_conditions}
q_\theta\big(y_{\tiny \mbox{test}} ~|~\{\bz_{i}, y_i \}_{i=1}^n, \bz_{\tiny \mbox{test}} \big) - \mathbb{E}_{\calD_n \sim p_n }\left[ q_\theta\big(y_{\tiny \mbox{test}} ~|~\{\bz_{i}, y_i \}_{i=1}^n, \bz_{\tiny \mbox{test}} \big) \right] & \rightarrow & 0 ~~~\mbox{(variance)} \nonumber \\
 p\big( y_{\tiny \mbox{test}} ~|~  \bz_{\tiny \mbox{test}} \big) - \mathbb{E}_{\calD_n \sim p_n }\left[ q_\theta\big(y_{\tiny \mbox{test}} ~|~\{\bz_{i}, y_i \}_{i=1}^n, \bz_{\tiny \mbox{test}} \big) \right] & \rightarrow & 0 ~~~\mbox{(bias)}
\end{eqnarray}
almost surely as as $n \rightarrow \infty$.  We are assuming here that $\bz  = g_\phi\big( \calG_H(\bx ) \big)$ and that $\{ p_n \}_{n=m}^\infty$ represents a sequence of RDB distributions adhering to the generative process from Definition \ref{def:reverse_time}; likewise for $p\big( y_{\tiny \mbox{test}} |  \bz_{\tiny \mbox{test}} \big)$ which does not vary with $n$.  Building on the rich conceptual framework from \citet{nagler2023statistical}, originally designed to analyze IID scenarios over single tables, we now examine\textit{ necessary} conditions on a given $\{q_\theta,g_\phi \}$ for achieving (\ref{eq:ideal_asymptotic_conditions}), noting that $g_{\tiny \mbox{pf}}$ is a special case of $g_\phi$.

\vspace*{0.1cm}
\begin{proposition} \label{prop:locality_stuff}
Given an RDB FM defined by some $\{q_\theta, g_\phi \}$, and every $\rho \in \calP$ adhering to Definition \ref{def:reverse_time}, assume (\ref{eq:ideal_asymptotic_conditions}) holds and that $p\big(y_i | \calG_H(\bx_i) \big) = p\big(y_i | \bz_i \big)$ $\forall i=0,\ldots,n$.  Then for any $\check{\rho} \in \calP$ (and associated label generator $\check{p}$) there exists a sequence $\varepsilon_n \rightarrow 0$ such that almost surely
\begin{equation} \label{eq:local_error}
 \big| q_\theta\big(y_{\tiny \mbox{test}} ~|~\{\bz_{i}, y_i \}_{i=1}^n, \bz_{\tiny \mbox{test}} \big) - q_\theta\big(y_{\tiny \mbox{test}} ~|~\{\check{\bz}_i, \check{y}_i \}_{i=1}^n, \bz_{\tiny \mbox{test}} \big)\big| \xrightarrow{n \rightarrow \infty} 0,
\end{equation}
where samples $\{\bz_i, y_i \}_{i=1}^n$ are extracted from $\calD_n \sim p_n$,  samples $\{\check{\bz}_i, \check{y}_i \}_{i=1}^n$ from $\check{\calD}_n \sim \check{p}_n$, and $p\big( y ~|~  \bz \big) = \check{p}\big( y ~|~  \bz \big)$ ~if~ $\|\bz - \bz_{\tiny \mbox{test}} \| \leq \varepsilon_n$.
\end{proposition}
\vspace{0.0cm}

Intuitively, this result merely indicates that if we wish to obtain unbiased estimates of the label across broad distributional settings within $\calP$ (as required by RDB foundation models in particular), then the output of our decoder function $q_\theta$ should depend only on ICL samples with encoder representations $\bz_i$ that are \textit{close} to the test representation $\bz_{\tiny \mbox{test}}$.  Note that 
\begin{equation}
\{\bz_i, y_i \}_{i=1}^n  =  \{\bz_i, y_i \}_{\bz_i \in \Omega_n}  \cup \{\bz_i, y_i \}_{\bz_i \not\in \Omega_n} ~~\mbox{with}~~\Omega_n = \{\bz : \|\bz - \bz_{\tiny \mbox{test}} \| \leq \varepsilon_n \},
\end{equation}
while $\{\check{\bz}_i, \check{y}_i \}_{i=1}^n$ decomposes analogously.  When $n$ grows large, elements of the sets $\{\bz_i, y_i \}_{\bz_i \in \Omega_n}$ and $\{\check{\bz}_i, \check{y}_i \}_{\check{\bz}_i \in \Omega_n}$ become indistinguishable as $\Omega_n$ collapses around $\bz_{\tiny \mbox{test}}$, hence they may influence the output of $q_\theta$ without conflicting with (\ref{eq:local_error}).  In contrast, remaining samples not in $\Omega_n$ must have negligible influence for (\ref{eq:local_error}) to hold.  This dependency on \textit{locality} allows the FM to differentiate RDBs that have similar behavior around any given test point, but increasing separation elsewhere.  We next discuss the implications for RDB FM design types, particularly those based on TFMs.

\vspace*{-0.2cm}
\subsection{Natural Alignment with Single-Table FMs.} \label{sec:ppd_stuff}
\vspace*{-0.1cm}

\paragraph{Uniform ICL sampling.} When choosing ICL samples, we would like at least some associated with encoder representations $\{\bz_i\}_{i=1}^n$ that are reasonably \textit{close} to the test point representation $\bz_{\tiny \mbox{test}}$, as required for unbiased estimation per Proposition \ref{prop:locality_stuff}.  In the absence of domain-specific knowledge or strong a priori assumptions such as homophily,\footnote{In the context of heterogeneous graphs or subgraphs defining RDB relations, \textit{homophily} refers to the tendency of neighboring nodes of the same node type (not necessarily 1-hop neighbors) to share the same label, while \textit{heterophily} represents the converse.} we may not know in advance where these close samples will be located in a given RDB.  And yet FM architectures should ideally address wide-ranging conditions with a shared data-agnostic strategy.  Along these lines, uniform sampling (from rows of $\bX$ with historical timestamps) is reasonable for approximating the full RDB context across generic new tasks, leading to at least some coverage of any new $\bz_{\tiny \mbox{test}}$ with select elements of $\{\bz_i\}_{i=1}^n$.\footnote{By uniform sampling, we do not exclude temporal bias.  For example, if we wish to prioritize recency, we can still uniformly sample across all RDB neighborhoods satisfying a time difference constraint.}  Moreover, in a sufficiently large RDB these uniform samples may even behave roughly as though they were IID, mirroring the single-table distributions for which TFMs were originally pre-trained.  This alignment can play a critical role in achieving more robust performance as the number of ICL samples grows, which we will investigate next through the lens of \textit{posterior predictive distributions} (PPDs) \citep{muller2021transformers}.  The PPD serves a critical role, representing the target distribution that ICL-based TFMs attempt to approximate during pre-training \citep{hollmann2022tabpfn}.

\vspace*{-0.1cm}
\paragraph{Leveraging Single-Table PPDs.}  Given some RDB $\calD$ drawn from ground-truth distribution $p_{\tiny \mbox{gt}}$, suppose we extract samples $\{\calG_H(\bx_i), y_i \}_{i=1}^n$ and test point $\{\calG_H(\bx_{\tiny \mbox{test}}), y_{\tiny \mbox{test}}\} = \{\calG_H(\bx_0), y_0\}$ satisfying
\vspace*{-0.15cm}
\begin{equation} \label{eq:factorized_rdb}
    p_{\tiny \mbox{gt}}\big(\{\calG_H(\bx_i), y_i \}_{i=0}^{n} \big) \approx \prod_{i=0}^{n} p_{\tiny \mbox{gt}}\big(\calG_H(\bx_i), y_i \big) ~~\mbox{and}~~  p_{\tiny \mbox{gt}}\big(y_i | \calG_H(\bx_i) \big) \approx p_{\tiny \mbox{gt}}\big(y_i | \bz_i) \big)~~\forall i.
    \vspace*{-0.15cm}
\end{equation}
These may be reasonable approximations if (i) uniform sampling is conducted over a sufficiently large RDB, and (ii) the embeddings $\bz_i$ adequately reflect the dominant sources of discriminative information within $H$-hop neighborhoods.  Related assumptions also appear in the graph learning literature \citep{ma2021subgroup}.  Granting the approximations from (\ref{eq:factorized_rdb}) hold as $n$ grows large, we may directly borrow an existing prior $\mu$ over \textit{single-table} distributions $q \in \calQ$ satisfying mild regularity conditions \citep{nagler2023statistical}.  We then form a PPD that  converges towards the ground-truth as
\vspace*{-0.0cm}
\begin{equation} \label{eq:ppd2}
p_{\tiny \mbox{ppd}}\big(y_{\tiny \mbox{test}} |\bz_{\tiny \mbox{test}}, \calS_n \big) ~ = ~ \int q \big( y_{\tiny \mbox{test}} | \bz_{\tiny \mbox{test}}  \big) d \mu \big( q | \calS_n \big)  ~~ \xrightarrow{n \rightarrow \infty} ~~ \hat{p}\big(y_{\tiny \mbox{test}} | \bz_{\tiny \mbox{test}} \big) ~\approx ~ p_{\tiny \mbox{gt}}\big(y_{\tiny \mbox{test}} | \bz_{\tiny \mbox{test}} \big),
\end{equation}
where $\hat{p}$ is the KL-optimal approximation of $p_{\tiny \mbox{gt}}$ in $\calQ$.
This drift towards the ground-truth $p_{\tiny \mbox{gt}}$ is essential for the following reason. While the distribution of pre-training data for TFMs may loosely resemble trends in real-world tables, there is certainly no exact correspondence.  However, a perfect match with the prior is \textit{not} required to achieve strong performance, provided that: (i) the prior assigns at least some nonzero probability mass to a $\hat{p}$ reasonably near $p_{\tiny \mbox{gt}}$, and (ii) we have collected enough uniformly-distributed ICL samples such that the true PPD narrows as 
$p_{\tiny \mbox{ppd}}\big(y_{\tiny \mbox{test}} |\bz_{\tiny \mbox{test}}, \calS_n \big) \approx \hat{p}\big(y_{\tiny \mbox{test}} | \bz_{\tiny \mbox{test}} \big)$.  Originally motivated from single-tables \citep{nagler2023statistical}, this phenomenon  may at least partially explain the observed empirical success generalizing to RDBs, whereby conditions mirroring the single-table case have been explicitly engineered through uniform sampling and encoders $g_{\tiny \mbox{pf}}$ that do not mix across columns, preserving column identity and roles akin to typical tabular regimes \citep{JUICE_paper_2026}.  The latter plausibly increases the chances that condition (i) from above is roughly satisfied in regions around $\bz_{\tiny \mbox{test}}$.  In so doing we effectively align RDB neighborhoods with attributes of input data previously applied to successful TFMs, transferring similar efficacy.

\vspace*{-0.1cm}
\subsection{Limitations of Alternative Possibilities}  \label{sec:random_walk_sampling}

\vspace*{-0.1cm}

\paragraph{Bias from non-uniform sampling.}  The distance between local RDB neighborhoods, as arbitrated by encoder representations $\{\bz_i\}_{i=1}^n$, determines the notion of locality upon which Proposition \ref{prop:locality_stuff} is based.  But it is nonetheless useful to contrast this notion with distinct options adopted by other types of RDB FMs.  For example, the latest SOTA relational Transformer instantiation RT-J \citep{ranjan2026rt,ranjanlarge}, based on RDB-specific pre-training with real-world RDB data, involves forming ICL samples as the set of target rows in $\bT^1$ that are most frequently visited over the course of a large number of fixed-length \textit{random walks} seeded from $\bx_{\tiny \mbox{test}}$.  In this way we collect samples that are close to $\bx_{\tiny \mbox{test}}$ in the sense of graph distance, \textit{but not necessarily in terms of encoder representation}.  For networks dominated by homophily effects, meaning nearby nodes in the graph have similar labels, this likely works well.  But more broadly, within an RDB influenced by heterophily relationships, disconnected regions, or other spurious network effects, similar samples/nodes drawn from random walks may \textit{not} be similar in terms of encoder representations, in which case the necessary conditions of Proposition \ref{prop:locality_stuff} would not hold.  This may also partially explain the performance of RT-J in Figure \ref{fig:single_table}, where the infused relational structure is minimally related to the labels by experimental design.  See also Appendix \ref{app:ICL_sampling_stuff} where we empirically show that RDBLearn with uniform sampling can achieve much \textit{lower} sample complexity relative to RW-based models (both RT-J and RDBLearn RW variants).  This implies that higher accuracy is possible even with fewer per-task context labels.

\vspace*{-0.2cm}

\paragraph{Implications for the PPD.} The distinction between uniform and non-uniform sampling also influences the PPD upon which ICL-based FMs depend.  For example, regardless of whether the measure $\mu$ reflects a single-table or a full RDB pre-training prior, if random-walk sampling is adopted to obtain $\calS_n$, then the core IID assumption from (\ref{eq:factorized_rdb}) used to establish (\ref{eq:ppd2}) will no longer generally hold.  Consequently, increasing the number of ICL samples need not approximate a desirable asymptotic regime, and the resulting PPD may bear little resemblance to the target $p_{\tiny \mbox{gt}}\big(y_{\tiny \mbox{test}} | \bz_{\tiny \mbox{test}} \big)$.  See Appendix \ref{app:random_walk_example} for a simple example whereby random-walks stall around the test node, failing to capture samples that lie elsewhere in the RDB as required to produce a meaningful PPD (the core issue is that $\calG$ is disconnected). Hence instead of having $n$ approximately IID samples from some $p_{\tiny \mbox{gt}}(\calG_H(\bx),y)$ (as with uniform sampling matched with a single-table prior $\mu$), with random walk sampling we instead have just \textit{one sample} from a much more complex, entangled joint distribution $p_{\tiny \mbox{gt}}\big(\{\calG_H(\bx_0), y_0, \},\ldots,  \{\calG_H(\bx_n),y_n \} \big)$.  This dramatically increases the necessity of an accurate \textit{RDB-specific} pre-training data prior to resolve ambiguity in the resulting PPD, as (\ref{eq:ppd2}) need not hold any longer, even approximately.  But such priors raise a new issue.

\vspace*{-0.2cm}

\paragraph{Dilemma of RDB-specific pre-training priors.}  Now suppose that $\mu$ is no longer simply a prior over distributions of single-tables with IID rows (as assumed by TFMs), but instead over distributions of full RDBs as adopted by multiple RDB FMs \citep{chen2026openrfm,fey2025kumorfm,hudovernik2026kumorfm,ranjan2026rt}.  But designing an accurate prior of this type is more challenging than in the single-table case, simply because of all the new relational dependencies introduced over multiple tables.  This implies that misspecifications may actually be \textit{greater} relative to real-world RDBs, and yet we cannot necessarily compensate with a larger $n$ (as in TFM cases) if we no longer induce an IID regime or the equivalent of (\ref{eq:ppd2}).  These observations suggest that, outside of specific domains where developing targeted pre-training priors may be more feasible, generic TFM-based pipelines still hold some sway.  Plausibly, loose IID approximations with even modest compensatory effects as $n$ grows large might often generalize better in new domains than highly-complex, entangled priors that do not factorize over ICL sample sets $\big( \{\calG_H(\bx_0), y_0, \},\ldots,  \{\calG_H(\bx_n),y_n \} \big)$.

\vspace*{-0.1cm}
\section{Adequate Encoder Expressivity even with No parameters}\label{sec:adequate_encoder}
\vspace*{-0.1cm}

We next address possible expressivity limitations associated with $g_{\tiny \mbox{pf}}$ usage, the key ingredient which allows TFMs to avoid any RDB-specific training or pre-training.  A natural presumption here is that parameter-free encoders place a potentially-burdensome restriction on the types of RDB neighborhood representations available for constructing ICL samples.  We analyze this concern through a case study involving interaction terms, where we find that the added expressivity of parameterized encoders $g_\phi$ need not provide an unequivocal advantage over the  much simpler $g_{\tiny \mbox{pf}}$ adopted by TFMs.

\vspace*{-0.2cm}

\paragraph{Interaction terms as a case study.} Prior work has noted that parameter-free encoders that process RDB columns individually (as mentioned in Section \ref{sec:experiments} and adopted by RDBLearn/TabPFN-Rel) are incapable of handling interaction terms between columns in auxiliary tables.  Along these lines, consider the following representative example of this ostensible limitation discussed by \citet{hudovernik2026kumorfm,huang2026relagent}.  Suppose we are given an RDB with only two tables: A labeled target table $\bT^1 = \by$ and a child table $\bT^2 = [\ba, \bb]$ composed of two binary feature columns $\ba$ and $\bb$.  A PK associated with each row of $\bT^1$ is linked by FKs to one or more rows of $\bT^2$, where $\calR_i$ denotes the set of rows within $\bT^2$ that have FKs pointing to the PK associated with row $i$ in $\bT^1$. Ground-truth labels are then constructed to satisfy
\vspace*{-0.1cm}
\begin{equation} \label{eq:toy_label_equation}
 \hspace*{0.0cm}   y_i =  \max_{j \in \calR_i} \left[a_j \land b_j \right]~~~ \forall i. 
    \vspace*{-0.0cm}
\end{equation}
Clearly the element-wise interactions $a_j \land b_j$ determine the label, and so the argument goes, excluding such interactions is disproportionately problematic within any $g_{\tiny \mbox{pf}}$ that encodes columns separately as advocated by \citet{JUICE_paper_2026}.  We provide three counter-points to this concern:

\vspace*{-0.1cm}

\paragraph{\textit{All} lossy encoding involves negative counter-examples.}  The labels computed via (\ref{eq:toy_label_equation}) involve a binary function over a $2|\calR_i|$-dimensional input space, giving $2^{2|\calR_i|} = 4^{|\calR_i|}$ possible input configurations.  \textit{But this is just one possible function sharing this input range}.  Assuming no permutation invariance across rows, there are $2^{\left(4^{|\calR_i|}\right)}$ distinct binary functions expressible over the same input space. Without external knowledge mandating which function is correct, an encoder that fails to losslessly preserve the entire input range will make it impossible to guarantee that a subsequent decoder can determine the correct labeling rule; a single missing bit can flip any labeling rule leading to errors within an arbitrary downstream classifier.

More broadly, all lossy encoding, \textit{whether parameterized or not}, opens up a set of inevitable counter-examples, whereby learning the correct function is infeasible; see \citet{JUICE_paper_2026}[Proposition 4.2] as one representative scenario.  This remains true even if task-specific labels are included within $\calG_H(\bx)$; more on this in Appendix \ref{app:input_labels}.  Consequently, the primary differentiating consideration is not whether counter-examples exist; rather, what matters is the relative frequency with which encoder-specific counter-examples appear in real-world RDBs.\footnote{This is admittedly difficult to quantify, as we usually only have reported results on final predictive accuracy.}  In general, functional expressivity is likely more critical in \textit{supervised} scenarios, where trainable degrees-of-freedom can be mapped to task-specific encoders that mitigate task-specific counter-examples (e.g., a particular set of interaction terms).  But for ICL-based FMs restricted to a fixed data-agnostic form, other factors such as interpretability or sample complexity may matter more than parameterized expressivity per se.

\vspace*{-0.1cm}

\paragraph{Arbitrary interactions challenge \textit{any} fixed encoder.}  Even if we restrict label generation to functions of interaction terms similar to (\ref{eq:toy_label_equation}), parameterized encoders still do not have a clear-cut advantage in real-world settings.  This is because real-world auxiliary tables often have many columns.  Excluding all higher-order interactions, there will still exist a quadratically expanding number of candidate interaction terms as the number of columns grow.  Without knowing in advance which are important, lossy encoder compression, \textit{parameterized or not}, will create the potential for prediction errors (i.e., there will always be cases where needed interaction signal depends on missing information). Appendix \ref{app:input_labels} analyzes how historical labels within $\calG_H(\bx)$ cannot generally resolve this ambiguity under certain settings.

\vspace*{-0.2cm}

\paragraph{Parameter-free encoders can (sometimes) handle interactions.}  For arguments sake, the presumption thus far in this section has been that a column-wise parameter-free $g_{\tiny \mbox{pf}}$ cannot represent interaction terms.  But actually, this is not strictly true for the following two reasons.  First, models like RDBLearn or TabPFN-Rel can directly handle interactions between compressed representations of individual columns; these representations form additional columns within $\bZ \in \mathbb{R}^{n\times d_z}$, and TFMs can readily handle interactions between the constituent ICL features.  Secondly, with enough aggregations of a suitable type, lossless compression is possible to enable interactions, at least in principle.  For continuous aggregators \citep{corso2020principal}, the required cardinality is equal to the number of real-valued elements being aggregated (not surprising), noting that these aggregators need not be permutation invariant (e.g., they could be time-dependent, involve symmetry-breaking projections, etc., for resolving ordered column values).  But if arbitrary discontinuities are allowed (which is admittedly not practically feasible to implement), even just a single aggregation is technically capable of reconstructing the full set \citep{rubio2026fixed}; this conclusion follows from the Kolmogorov-Arnold representation theorem \citep{schmidt2021kolmogorov}.

\begin{figure*}[t!]
\centering
\includegraphics[height=3.5cm]{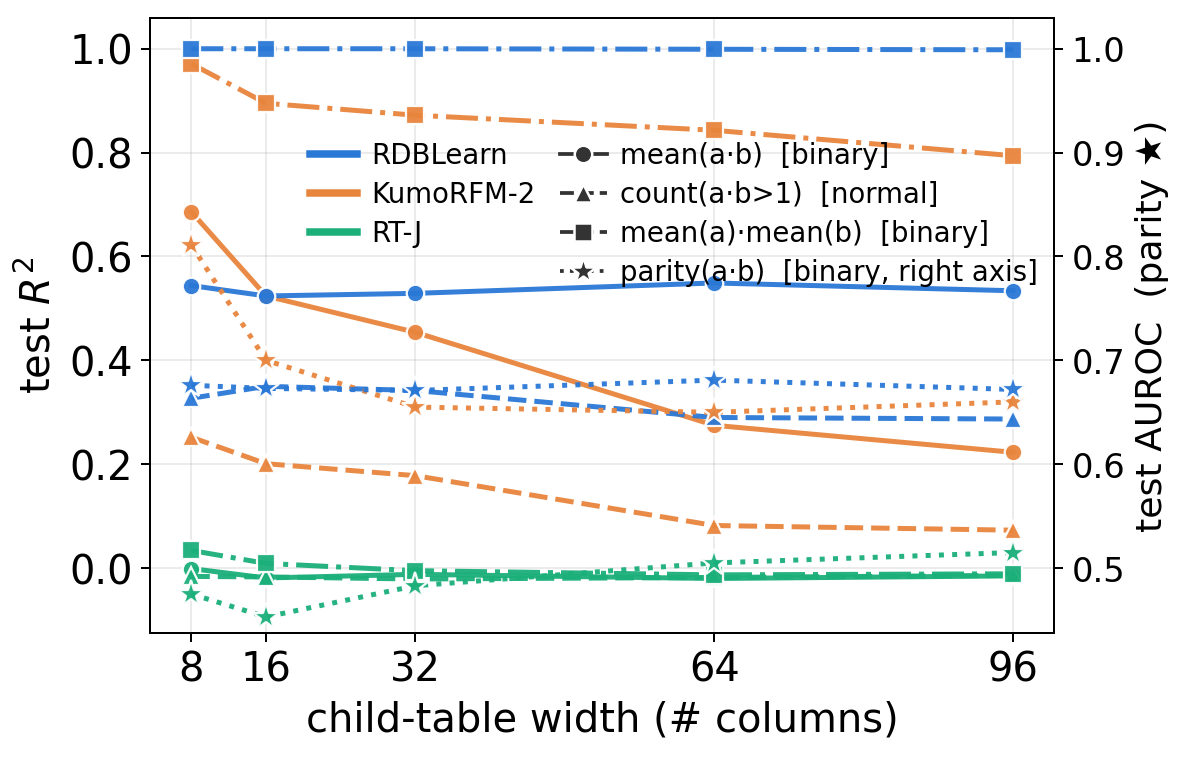}%
\includegraphics[height=3.6cm]{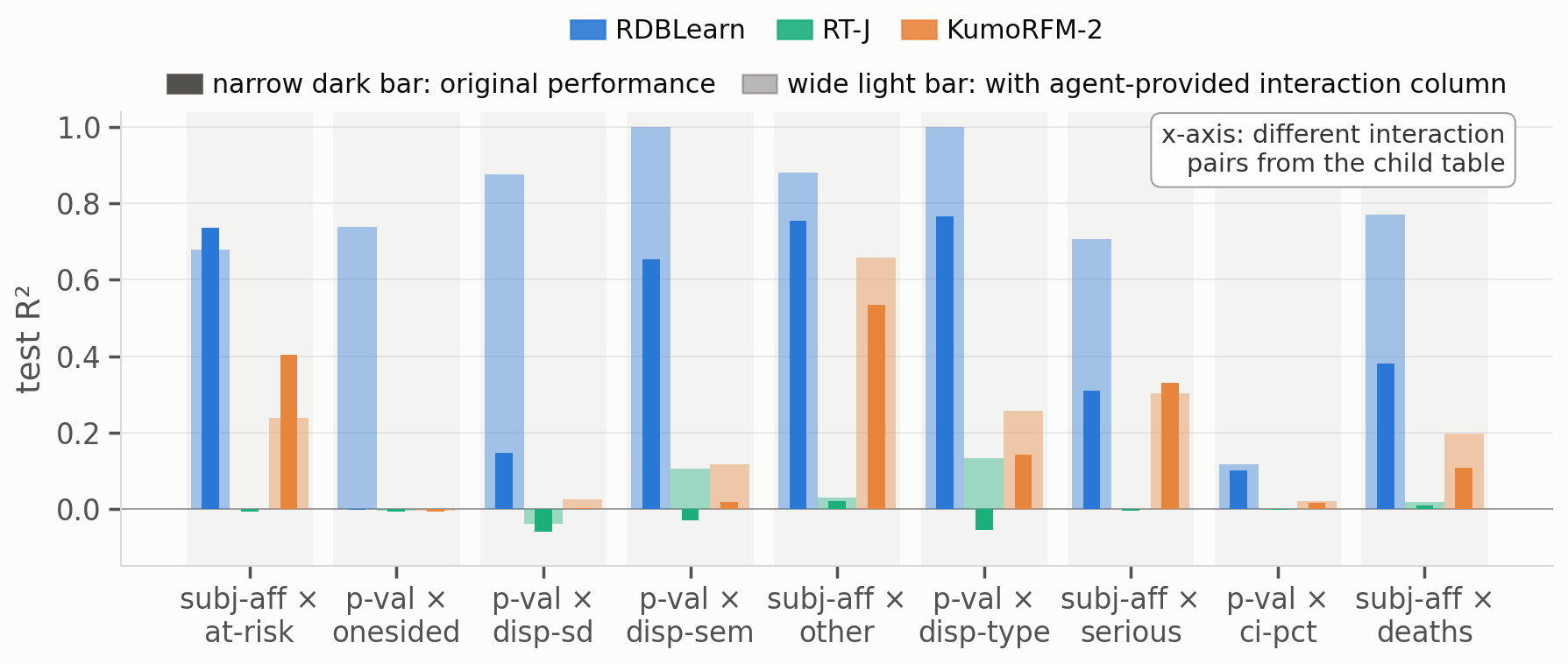}%
\vspace*{-0.1cm}
    \caption{\textit{Left}: Prediction quality over synthetic RDBs formed via 4 different interaction types in child table $\bT^2$.  Colors denote models; line styles distinguish the 4 RDB versions.  Results are shown as the width of $\bT^2$ increases by padding with uninformative columns.  \textit{Right}: Performance on quasi-synthetic data, whereby new ground-truth labels are generated via interactions between different column pairs (x-axis) within the real-world RDB rel-trial.  Solid/dark bars show original results, while shaded/light bars represent revised performance after appending needed interaction terms to the appropriate table. The latter simulates an agentic scenario described in Section \ref{sec:agents}.}
    \label{fig:interaction_full}
    \label{fig:interaction_tests}
    \label{fig:interations}
  \vspace*{-0.2cm}
\end{figure*}

\vspace*{-0.2cm}
\paragraph{Interaction experiments.}  We conduct multiple tests to explore the impact of interaction terms; see Appendix \ref{app:interaction_experiments} for full details.  For the first, we generalize the setup of (\ref{eq:toy_label_equation}) to include 4 different interaction types involving  columns $\{\ba, \bb\}$ in $\bT^2$; we then compare TFM-based RDBLearn against the top non-TFM-based RDB FMs, KumoRFM-2 and RT-J.  Figure \ref{fig:interaction_full} (\textit{left}) shows the prediction results as the width of $\bT^2$ is increased by adding uninformative columns.  These distracting columns follow the same distribution as $\ba$ and $\bb$, and models do not know in advance which columns are useful, raising estimation difficulty.  Overall, we observe that RT-J does not work well in these settings, likely because the implicit graph is disconnected, which can disrupt the collection of suitable ICL samples (see Section \ref{sec:random_walk_sampling}).  Meanwhile, RDBLearn is consistently competitive with KumoRFM-2, and actually displays more stable performance across table width. This occurs because the aggregated features from $g_{\tiny \mbox{pf}}$ can be combined by the subsequent TFM, at times forming usable approximations despite not explicitly computing most interaction terms (see further details in Appendix \ref{app:interaction_experiments}).  Although we do not pursue further here, expanding the aggregation set within $g_{\tiny \mbox{pf}}$ can in principle reduce RDBLearn prediction errors beyond what is shown.

For the second test, we begin with the rel-trial dataset \citep{robinson2024relbench} and compute new ground-truth labels via interactions between sets of auxiliary table columns.  Results are shown in Figure \ref{fig:interaction_full} (\textit{right}) across 9 different column interaction pairs (x-axis) used for generating labels. While these tasks are challenging for all methods, RDBLearn generally outperforms the others (agentic results included on this bar plot are addressed in Section \ref{sec:agents}).

\vspace*{-0.1cm}
\section{TFMs Form Natural Partnering with Agentic Systems} \label{sec:agents}

\vspace*{-0.1cm}

\paragraph{Feature discrimination in three stages.} RDB predictive modeling depends heavily on the ability to differentiate, on a task-specific basis, informative versus non-informative features, relational or otherwise.  There are (at least) three candidate stages where such feature screening may plausibly take place: symbolically prior to any FM application, via \textit{reasoning agents} with access to task-dependent meta-data or domain expertise (\textit{stage 1}); locally at the RDB encoder level using whatever context is available within $\calG_H(\bx)$ (\textit{stage 2}); and globally at the decoder level in a purely data-driven manner, using the full range of labeled ICL samples (\textit{stage 3}).  It is our assertion that TFM-based models are particularly adept at utilizing all three stages as \emph{complements}, particularly because their $g_{\tiny \mbox{pf}}$ can reduce potential interference between stages.  In support of this conjecture, we offer the following:

\vspace*{-0.3cm}

\paragraph{Stage 2 as a conduit from the symbolic to the data-driven.}  Given the previously stated capabilities of stages 1 and 3, a plausible role for stage 2 becomes relatively transparent: Make sure any features that stage 1 judges to have potential are organized, to the extent possible, so as to align with the distribution of single-table features for which the subsequent stage 3 ICL decoder has been optimized.  Among other things, this entails that if a particular set of features are designated as possibly useful by stage 1, they are not mixed together; some may be valuable, others ultimately not, but that is for stage 3 to decide in a data-driven manner (per results from \citet{JUICE_paper_2026}, this task is often easier when column-wise axis-alignment is maintained).  Most notably, a so-called ``golden feature'' with designated predictive value by a stage 1 agent should be passed directly through stage 2 without alteration or attenuation.  For example, this feature could be a specific interaction term whose \textit{a priori} importance would be otherwise indeterminate without domain knowledge from stage 1.  Or alternatively, various aggregations over temporal windows might be incorporated to reflect known recency biases.  Of course numerous other pragmatic possibilities exist as well, which can be iteratively explored by an agentic system, with robust and scalable validation facilitated by TFMs.

\vspace*{-0.3cm}

\paragraph{TFMs innately honor stage 1--3 roles.}  When we adopt TFMs, equipped with an appropriate $g_{\tiny \mbox{pf}}$, any designated golden feature from stage 1 can simply be appended to others in stage 2 without deviating from the original tabular setting under which stage 3 pre-training was conducted.  But this need not be the case for existing RDB FMs with  parameterized $g_\phi$.  For the latter, there is no guarantee that any input dimension/feature, golden or not, will be preserved within the output embedding.  Moreover, if we simply append such features to the encoder output (bypassing the parameterized encoder itself), we then introduce potential mismatch from the original RDB-specific pre-training regime of these architectures.  There is also the larger related issue of \textit{interpretability}. An agent can directly reason over the SQL-based aggregations used to form $g_{\tiny \mbox{pf}}$ and offer suggestions regarding what useful factors might be missing.  In contrast $g_\phi$ is much more opaque since the output embeddings result from stacked Transformer layers, e.g., there is no obvious way for an agent to tell if a valuable interaction term is attenuated or absent altogether.

\vspace*{-0.3cm}

\paragraph{Empirical corroboration.}  Consider the hypothetical scenario whereby an agent (stage 1) with domain knowledge materializes a golden feature; namely, an interaction term which determines the downstream label is added as a new column (see Appendix \ref{app:interaction_experiments}).  Figure \ref{fig:interaction_full} (\textit{right}) results show that RDBLearn leverages this information to consistency boost performance in line with the three-stage perspective from above.  RT-J and KumoRFM-2 experience more modest gains, and in certain cases do not capitalize at all on the provided golden feature (e.g., p-val$\times$onesided, p-val$\times$disp-sd cases from the figure), suggesting that the respective pipelines have somehow attenuated or underutilized this golden feature.  We further explore real-world implications in Appendix \ref{app:agent_experiments}, whereby RDBLearn with a single coding-agent prompt can match the performance of a SOTA multi-agent system \citep{huang2026relagent} that trains numerous models \textit{supervised} on a per-task basis.  Moving forward, we also emphasize that a \textit{default} expectation should be that any agentic system built on top of tabular base models will likely exceed the benchmark performance of pipelines such as RDBLearn and TabPFN-Rel-OSS.  This is because as public open-source projects, their design choices  represent an observable starting point with known performance upon which to calibrate any additional enhancements, albeit at the cost of significant leakage concerns discussed in Appendix \ref{app:LLM_usage}.

\vspace*{-0.2cm}
\section{Take-Home Message}
\vspace*{-0.2cm}
Despite never seeing relational data (real or synthetic) during their pre-training stage, empirical results indicate that TFMs paired with a simple $g_{\tiny \mbox{pf}}$ are competitive with the best alternative RDB FM architectures.  Although our core analyses provide multiple complementary rationales for this phenomenon, we readily concede that advancements in the field may plausibly shift the balance moving forward, hopefully catalyzed in part by our findings herein. 

\vspace*{-0.2cm}

\bibliography{reference_from_muse,refs,wipf_refs,reference}
\bibliographystyle{arxiv2027}

\clearpage
\appendix

\section{Further Analysis of the Continuum Perspective  from Section \ref{sec:continuum}}

In this section we demonstrate that simple ensembling or interpolation strategies, designed to combine canonical TFMs (without $g_{\tiny \mbox{pf}}$) and RDB-native FMs, may struggle to handle data along a continuum between single-tables and RDBs.  Later, we discuss further benchmarking implications of the continuum perspective, before concluding with additional semi-synthetic experiments.

\subsection{Ensembling Different FM Types Cannot Resolve the Continuum} \label{app:ensembling_limitation}

One candidate workaround for navigating varying relational effects on a continuum would be to invoke both a typical RDB FM and a single-table FM, and then simply use the validation set to either pick the best one or form some type of ensemble.  To analyze this possibility, we define predictions from tabular and RDB FMs respectively as
\begin{eqnarray} \label{eq:tab_RDB_FMs}
  \hat{y}_{\tiny \mbox{test}}^{\tiny \mbox{tab}}  & := & f_\theta^{\tiny \mbox{tab}}\big(\{\bz_{i}, y_i \}_{i=1}^n, \bz_{\tiny \mbox{test}} \big) ~~\mbox{s.t.}~ \bz  =  \bx \\ 
  \hat{y}_{\tiny \mbox{test}}^{\tiny \mbox{rdb}} &  := & f_\theta^{\tiny \mbox{rdb}}\big(\{\bz_{i}, y_i \}_{i=1}^n, \bz_{\tiny \mbox{test}} \big) ~~\mbox{s.t.}~ \bz  = g_\phi\big( \calG_H(\bx ) \big) \nonumber
\end{eqnarray}
for an arbitrary input RDB.  The TFM from (\ref{eq:tab_RDB_FMs}) only has access to $\bT^1$ information, or any tables that can be trivially joined to $\bT^1$ without a true encoder $g_{\tiny \mbox{pf}}$ (i.e., by design we are assuming standard single-table operational mode here, not a TFM equipped with $g_{\tiny \mbox{pf}}$ for handling rich relational effects).  To formalize things further, we require precise specifications of a suitable RDB generation process.  We must also introduce some form of practical constraint on the RDB FM encoder; otherwise, it is trivially a superset of the TFM listed in (\ref{eq:tab_RDB_FMs}).  We present these definitions first, before proceeding to the main result of this section.

\vspace*{0.3cm}
\begin{definition} \label{def:RDB_gen_process}
    We define $\calD_\tau$ as an RDB induced by a single table via the following generation process $p_\tau$, where continuation parameter $\tau \in [0,1]$ mitigates the significance of cross-table relational factors.  First, we draw tabular rows $\bx_{i} = \left[ \bx_{i}^{\tiny \mbox{real}}; \bx_{i}^{\tiny \mbox{cat}} \right]$ from some $p(\bx_{i})$,  where $\bx_{i}^{\tiny \mbox{real}} \in \mathbb{R}^{d}$ represent numerical features, $\bx_{i}^{\tiny \mbox{cat}} \in \calC^{\kappa}$ are categorical features with $|\calC| \leq B$ and $i=0,\ldots, N$.  We assume that the marginal $p\left(\bx^{\tiny \mbox{real}}\right)$ is an absolutely continuous distribution satisfying $p\left(\bx^{\tiny \mbox{real}}\right) \leq p_{\max}$ for all $\bx^{\tiny \mbox{real}} \in \calR^d$.  We treat each categorical column as an FK pointing to a new auxiliary table $\bT^k$ (with $k \in \{2,\ldots,\kappa + 1 \}$) composed of a single PK column (the elements of which correspond with the unique categories in the referring FK column).  And finally, we generate labels by first uniformly sampling $d' < d$ indices from $\{1,\ldots,d\}$ to form the reduced numerical features $\widetilde{\bx}^{\tiny \mbox{real}}_{i} \in \mathbb{R}^{d'}$.  We then compute labels as
    \begin{equation} \label{eq:label_function2}
    y_i ~ = ~ \psi\big[ (1-\tau)f_1(\widetilde{\bx}^{\tiny \mbox{real}}_{i} ) + \tau f_2\big( \calG_H(\bx_{i}) \backslash \bx_{i} \big) \big] ~\in~\calY, ~~ \forall i = 0,\ldots,N,
\end{equation}
where $f_1$ and $f_2$ are continuous functions mapping their respective domains to a shared $r$-dimensional embedding space $\mathbb{R}^r$, after which $\psi : \mathbb{R}^r \rightarrow \calY$ converts to the label space.\footnote{For reference, $\calG_H(\bx) \backslash \bx$ represents all $H$-hop neighborhood information available at the timestamp associated with $\bx$ (following the newly-generated PK-FK pairs), but excluding the seed row features $\bx$.}  We  reserve $\{\calG_H(\bx_{i}), y_i \}_{i=1}^n$ for ICL, with $n < N$,\footnote{This is effectively equivalent to uniform sampling given the original construction with IID rows.} while treating sample $i=0$ as $\{\calG_H\left(\bx_{\tiny \mbox{test}} \right), y_{\tiny \mbox{test}}\}$.
\end{definition}

\vspace*{0.3cm}
\begin{definition} \label{def:compressed_embedding} 
We define a space of minimally constrained RDB encoders $\bar{g}$, elements of which produce fixed-length embeddings as $\bz  = g_\phi\big( \calG_H(\bx ) \big) \in \mathbb{R}^{d_z}$.  Furthermore, we impose that
\begin{equation} \label{eq:reconstruction_restriction}
     \inf_{g_\phi \in \bar{g}} \inf_\eta  \mathbb{E}_{\calD_\tau \sim p_\tau } \big[ \| \bx^{\tiny \mbox{real}}_{\tiny \mbox{test}}  - \eta\big(\{\bz_{i}, y_i \}_{i=1}^n, \bz_{\tiny \mbox{test}} \big) \|  \big] > 0,
    \end{equation}
with expectation over the generation process from Definition \ref{def:RDB_gen_process} and $\eta : \mathbb{R}^{d_z \times (n+1)} \times \calY^{n} \rightarrow \mathbb{R}^d$ as an arbitrary continuous mapping. 
\end{definition}
\vspace*{0.3cm}

The intuitive implication of Definition \ref{def:compressed_embedding} is simply that $\bx^{\tiny \mbox{real}}_{\tiny \mbox{test}}$, which may be high-dimensional, cannot be completely reconstructed (in expectation) using all of the encoded information available to an ICL-based decoder $q_\theta$ (namely $\{\bz_{i}, y_i \}_{i=1}^n$ and $\bz_{\tiny \mbox{test}}$), regardless of which $g_\phi$ is chosen.  This is nominally a reasonable restriction to impose given that typical RDB encoders must ultimately compress into fixed-length embeddings, and such compression may apply to an original feature row $\bx$,  particularly so for any high-dimensional numerical values.  Moreover, (\ref{eq:reconstruction_restriction}) does \textit{not} rule out complete reconstruction of $d-1$ elements of $\bx^{\tiny \mbox{real}}_{\tiny \mbox{test}}$, indicating minimal enforced compression.  Given these definitions, we now analyze the limitation of ensembling:

\vspace*{0.3cm}
\begin{proposition} \label{prop:interpolation_limitation}
For any $\varepsilon \in (0,1)$ and $C > 0$ there will always exist fixed mappings $\{\Psi',f_1',f_2'\}$ such that the following simultaneously hold:
\vspace*{-0.2cm}
\begin{eqnarray} \label{eq:prop_interpolation_conditions}
\min_{f_\theta^{\tiny \mbox{tab}}} \mathbb{E}_{\calD_\tau \sim p_\tau} \Big[~ \big| y_{\tiny \mbox{test}} -  \hat{y}_{\tiny \mbox{test}}^{\tiny \mbox{tab}}  \big|~ \Big] = 0 & \tau = 0 \nonumber \\
\min_{f_\theta^{\tiny \mbox{rdb}},~ g_\phi \in \bar{g}}\mathbb{E}_{\calD_\tau \sim p_\tau} \Big[~ \big| y_{\tiny \mbox{test}} -  \hat{y}_{\tiny \mbox{test}}^{\tiny \mbox{rdb}}  \big|~ \Big] = 0 & \tau = 1 \\
    \min_{f_\theta^{\tiny \mbox{tab}},~f_\theta^{\tiny \mbox{rdb}},~g_\phi \in \bar{g},~\alpha \in [0,1]} \mathbb{E}_{\calD_\tau \sim p_\tau} \Big[~ \big| y_{\tiny \mbox{test}} - \big( \alpha \hspace*{0.05cm} \hat{y}_{\tiny \mbox{test}}^{\tiny \mbox{tab}} + (1-\alpha) \hat{y}_{\tiny \mbox{test}}^{\tiny \mbox{rdb}}\big)  \big|~ \Big] \geq C & \hspace*{0.3cm} \forall ~ \tau \in [\varepsilon,1-\varepsilon], \nonumber
\end{eqnarray}
where $\calD_\tau \sim p_\tau$ indicates an RDB sampled according to Definition \ref{def:RDB_gen_process} with $\{\Psi,f_1,f_2\} = \{\Psi',f_1',f_2'\}$ and $n > d'$, and $\bar{g}$ denotes any RDB encoder satisfying the conditions of Definition \ref{def:compressed_embedding}. 

\end{proposition}

Even though we are optimizing separately for each $\tau$, this relatively straightforward result shows that any additive combination of the two FMs (as produced by typical ensembling) can perform arbitrarily badly.  This is despite the fact that for the purely tabular regime ($\tau = 0$), the TFM exhibits ideal behavior, while for the purely relational case (i.e., $\tau = 1$ implying no dependence on the original target row features) the RDB FM performs perfectly.

Admittedly, we can always try training a more complex combining function to address this scenario, but then we are diverging from the targeted FM regime to supervised learning.  In contrast, if we pre-train our original FMs to consistently handle such cases, then in principle we should arrive at a model the operates effectively on both single and multiple tables.  \textit{But of course this is the main point we are advocating for in the first place; namely, RDB FMs should work on single tables or single-table adjacent regimes as well}.

\subsection{Benchmarking Implications of the Continuum Perspective} \label{app:benchmarking_implications}

If there were always a clear-cut distinction between single tables and RDBs, then benchmarking would be straightforward: simply partition into separate domains and evaluate accordingly, RDB FMs tested only on RDBs (e.g., RelArena, RelBench, 4DBInfer, etc.) and TFMs only on single tables (e.g., TabArena).  But in light of the continuum described in Section \ref{sec:continuum}, is such strict partitioning unequivocally warranted?  

For example, consider breaking out of this mold by applying an RDB FM to a suite of traditional tabular benchmarks.  If performance is competitive relative to frontier TFMs that specialize on single tables, then our confidence in the robustness of this RDB FM justifiable grows.  But now suppose instead that performance is notably lagging.  It is tempting to simply discount such cases as the magisteria of single-table FMs and move on; however, the continuum perspective complicates any such immediate dismissal.  After all, if an RDB FM does not work well on single-tables, what about performance on RDBs along a continuum that are minimally different from single tables?  More broadly, it is reasonable to question how large is the multi-table relational neighborhood around any single-table benchmark within which unsatisfactory predictions are likely to be produced.  Or conversely, we may ask how large along any continuum must this neighborhood grow before the RDB FM capabilities begin to outshine TFM alternatives by exploiting relational effects.  It is precisely these factors that (at least partially) motivate RDB FMs based on TFMs that already perform well on single tables.

\subsection{Semi-Synthetic Continuum Experiments} \label{app:semi_syn_experiments}

In Section \ref{sec:continuum} experiments we relied on real-world labeled data curated with the expectation that relational effects were nominally minimal. We now present complementary experiments involving new semi-synthetic data, whereby original labels are explicitly modulated by relational effects injected through (\ref{eq:label_function}); everything else follows the setup of Figure \ref{fig:single_table}. To instantiate this procedure over varying $\tau$, we implement (\ref{eq:label_function}) with $f_1$ equal to original/real data set labels and $\Psi$ set to a leaky-ReLU function.  Lastly, we form $f_2$ by collecting 2-hop neighbors over the induced graph, computing column skewness features for each neighborhood, concatenating, and then passing through a randomized MLP.  In this way, $\tau = 0$ is equivalent to the original data label subject to an additional non-linearity, while $\tau = 1$ is independent of original labels and entirely determined by induced relational effects.  

We execute this process across all regression tasks associated with Figure \ref{fig:single_table}, and repeat for different values of $\tau \in [0,1]$.  Table \ref{tab:vary_tau} shows $R^2$ values averaged over all tasks at each $\tau$, noting that RDBLearn is run with TabPFN-v3 as the sole base TFM for direct comparison purposes.  From these results, we observe that RDBLearn achieves the best performance across the tested spectrum, while the second best flips from TabPFN-v3 (w/ smaller $\tau$) to KumoRFM-2 (w/ larger $\tau$).  Although TabPFN-v3 performance falls off for large $\tau$, becoming worse than all RDB FMs, this is expected since we have not granted this model any relational features.

\begin{table}[h]
\small
\centering
\setlength{\tabcolsep}{5.5pt}
\renewcommand{\arraystretch}{1.15}
\caption{Performance on semi-synthetic regression tasks as $\tau$ is varied following the continuum perspective.  Reported numbers represent $R^2$ values (higher is better) averaged across all tasks.  Bold entries are best per row, underlined implies second best.}
\label{tab:vary_tau}
\begin{tabular}{c|cccc}
\toprule
$\tau$ & TabPFN-v3 & RT-J & KumoRFM & RDBLearn \\
\midrule
0.00 & \underline{0.351} & 0.065 & 0.269 & \textbf{0.368} \\
0.25 & \underline{0.300} & 0.088 & 0.222 & \textbf{0.415} \\
0.50 & 0.036 & 0.061 & \underline{0.203} & \textbf{0.430} \\
0.75 & 0.027 & 0.040 & \underline{0.187} & \textbf{0.431} \\
1.00 & -0.264 & 0.031 & \underline{0.252} & \textbf{0.507} \\
\bottomrule
\end{tabular}
\end{table}


\section{Implications of Encoder Label Usage}\label{app:input_labels}

Two counter-points challenge the efficacy of the parameter-free encoders $g_{\tiny \mbox{pf}}$  upon which TFMs applied to RDBs are based.  Both involve the use of observable labels as inputs to a trainable $g_\phi$, and the potential benefits of doing.  We analyze these scenarios next, providing new insights that further inform foundational questions of RDB FM development.

\subsection{Labels as Extra Discriminative Features} \label{app:labels_as_features}

Per previously-stated definitions, an RDB neighborhood subgraph $\calG_H(\bx)$ may include nodes associated with other rows of $\bT^1 \equiv \bX$, provided these rows are reachable within $H$ PK-FK hops and have an earlier time-stamp to avoid temporal leakage.  As these rows have associated labels, it is natural to consider including them as extra node features to improve downstream predictability.  We adopt $\calG_H^*(\bx)$ to reference the resulting subgraph with these labels concatenated as additional node features; we are assuming here that any labels included within $\calG^*_H(\bx)$ have time-stamps earlier than $\bx$. In more traditional supervised learning regimes involving graphs, this is a common practice for improving node classification accuracy \citep{wang2022propagate} (akin to a trainable form of label propagation).  More recently, analogous ideas have been applied to RDB foundation models, whereby trainable encoders are explicitly designed to process past labels of similar entities within an RDB neighborhood \citep{chen2026openrfm,hudovernik2026kumorfm,ranjan2026relational}.

But to what degree can we actually rely on such neighborhood labels as predictive features, specifically in the context of trainable RDB foundation model encoders $g_\phi$ applied to $\calG_H^*(\bx_{\tiny \mbox{test}})$ evaluated at an unlabeled test row of interest $\bx_{\tiny \mbox{test}}$. To begin, for an arbitrary RDB we define $\calG^*_{\tiny \mbox{test}}$ as an augmented version of $\calG$, with $\bx_{\tiny \mbox{test}}$ appended to $\bX$, known labels associated with $\bX$ included as candidate node features, but restricted to entities with timestamps earlier than $\bx_{\tiny \mbox{test}}$.  We also let $\widetilde{\calG}^*_H(\bx_{\tiny \mbox{test}})$ denote $\calG^*_{\tiny \mbox{test}} \backslash \calG^*_H(\bx_{\tiny \mbox{test}})$, meaning all RDB context outside of the label-infused $H$-hop ego-network centered at test point $\bx_{\tiny \mbox{test}}$. And lastly, we adopt $y_H(\bx_{\tiny \mbox{test}})$ to indicate the set of all past labels present within $\calG^*_H(\bx_{\tiny \mbox{test}})$; this of course does \textit{not} include the unknown label $y_{\tiny \mbox{test}}$ associated with $\bx_{\tiny \mbox{test}}$ itself.  We then have the following:

\begin{proposition}\label{prop:labels_as_features}
  Given any ego-network $\calG^*_H(\bx_{\tiny \mbox{test}})$ with $|y_H(\bx_{\tiny \mbox{test}})| > 0$, distributions over remaining RDB content $p\big(\widetilde{\calG}^*_H(\bx_{\tiny \mbox{test}}), y_{\tiny \mbox{test}} | \calG^*_H(\bx_{\tiny \mbox{test}}) \big)$ will always exist such that:
  \vspace*{-0.2cm}
    \begin{enumerate}
        \item $\mathrm{(Foundation~model~setting)}$~ $y_{\tiny \mbox{test}}$ is a deterministic function of $\calG^*_{\tiny \mbox{test}}$,  and yet any possible fixed encoder $g_\phi$ and prediction head $q_\theta$ perform no better than chance estimating $y_{\tiny \mbox{test}}$ in the sense that 

\begin{equation}
\mathbb{E} \left[ \mathbb{KL}\Big[ p(y_{\tiny \mbox{test}} |\calG^*_{\tiny \mbox{test}} )  ~||~ q_\theta\big(y_{\tiny \mbox{test}} ~|~ g_\phi\big[ \calG^*_H(\bx_{\tiny \mbox{test}}) \big]  \big) \Big] \right]  ~~  \geq ~~ \mathbb{E} \left[\mathbb{KL}\Big[ p(y_{\tiny \mbox{test}} |\calG^*_{\tiny \mbox{test}} )  ~||~ \mathrm{Unif}\big( y_{\tiny \mbox{test}} \big) \Big] \right], 
\end{equation}
with outer expectations over $p\big(\widetilde{\calG}^*_H(\bx_{\tiny \mbox{test}}) | \calG^*_H(\bx_{\tiny \mbox{test}}) \big)$.

    \item $\mathrm{(Supervised~setting)}$~ Given $\{\calG^*_H(\bx_{i}), y_i \}_{i=1}^{n}$ and $y_{\tiny \mbox{test}}$ extracted from any single draw from $p\big(\widetilde{\calG}^*_H(\bx_{\tiny \mbox{test}}), y_{\tiny \mbox{test}} | \calG^*_H(\bx_{\tiny \mbox{test}}) \big)$, there exists a draw-specific $f_\theta = q_\theta \circ g_{\tiny \mbox{pf}}$ satisfying $y_i = f_\theta \big[\calG^*_H(\bx_{i}) \big]$ for all $i$ and $y_{\tiny \mbox{test}} = f_\theta \big[\calG^*_H(\bx_{\tiny \mbox{test}}) \big]$, where $g_{\tiny \mbox{pf}}$ is a parameter-free mapping from any $\calG^*_H(\bx_{\tiny \mbox{test}})$ to $\mathbb{R}^m$ and $q_\theta$ is a ReLU network with finite width and depth.
     \end{enumerate}
\end{proposition}

\textbf{Intuition behind Proposition \ref{prop:labels_as_features}.}  Neighborhood labels within $\calG_H^*(\bx)$ can be highly discriminative in a \textit{supervised} setting, where training specifically on samples $\{\calG^*_H(\bx_{i}), y_i \}_{i=1}^{n}$ extracted from a \textit{single}  RDB predictive task facilitates learning a task-specific discriminative function $f_\theta$ applicable to test points $\calG^*_H(\bx_{\tiny \mbox{test}})$ (part 2). And yet it is not generally possible to determine a stable predictive role for these labels within an encoder that must remain \textit{fixed} across varying datasets/tasks facing a foundation model architecture of the form  $q_\theta\big(y ~|~ g_\phi\big[ \calG^*_H(\bx_{\tiny \mbox{test}}) \big]  \big)$ (part 1). Given this implication of Proposition \ref{prop:labels_as_features}, why then are existing schema-agnostic foundation models like the original relational Transformer (RT) model from \citet{ranjan2026relational} capable of making successful predictions?  A simple plausible hypothesis is that, within the handful of real-world RDBs available for pre-training RT models, homophily relationships dominate, so simply adopting the majority label of local neighbors can be a highly discriminative feature.  Empirical results from \citet{ranjan2026relational}[Table 1] support this conclusion, namely, the similar performance of a basic entity mean predictor explicitly predicated on homophily (66.7 avg AUROC) relative to RT (69.7 avg AUROC).

\vspace*{-0.1cm}
 \paragraph{Implications for ICL-based models.}  ICL provides one pathway for diversifying label usage effectively beyond the homophily-dominant regimes mentioned above; however, this remains true whether or not a parameter-free encoder $g_{\tiny \mbox{pf}}$ is adopted. This is because even without trainable parameters, encoder representations can include diverse label aggregations that are sensitive to both homophily and heterophily relationships; the subsequent ICL-prediction head then sorts which are relevant for any given task.  Note that canonical aggregators can trivially handle the parity check example constructed within the proof of Proposition \ref{prop:labels_as_features} (see Appendix \ref{app:proofs}).  Hence given evidence available thus far, the use of labels as discriminative features does not by itself provide a compelling reason to favor trainable parameterized RDB encoders $g_\phi$ within ICL-based models.  And so TFMs equipped with $g_{\tiny \mbox{pf}}$ ultimately remain viable independently of whether or not such labels are available.

\subsection{Labels for Adjudicating Feature Importance}  \label{app:labels_and_feature_importance}

Analysis from \citet{JUICE_paper_2026} indicates that uninformative feature columns within RDBs cannot generally be filtered out using a dataset-agnostic encoder, which reduces the motivation for including trainable encoder parameters.  However, this calculus may change if the encoder is provided with local neighborhood labels (as in $\calG_H^*(\bx)$).  The latter provide task-specific information, which may plausibly facilitate feature-column differentiation on a task-by-task basis as required by RDB foundation models.  But is such differentiation generally possible even when $\calG_H(\bx)$ is switched to $\calG_H^*(\bx)$?  The answer is negative in the following sense:

\begin{proposition}\label{prop:labels_and_feature_importance}
  Given any representative RDB ego-network $\calG_H(\bx_{\tiny \mbox{test}})$ constructed from non-empty tabular feature columns and $|y_H(\bx_{\tiny \mbox{test}})| > 0$, there will always exist distributions over $\widetilde{\calG}^*_H(\bx_{\tiny \mbox{test}})$ and $y_H(\bx_{\tiny \mbox{test}}) \cup y_{\tiny \mbox{test}}$ such that:
  \vspace*{-0.2cm}
    \begin{enumerate}
        \item Each label in $y_H(\bx_{\tiny \mbox{test}}) \cup y_{\tiny \mbox{test}}$ is a shared deterministic function of some support set $S$ of tabular column features and independent of all remaining columns.

        \item Encoder input labels $y_H(\bx_{\tiny \mbox{test}})$ provide no information about the unknown generating support set $S$ in the sense that $p\big(S  | \calG^*_H(\bx_{\tiny \mbox{test}}) \big) = p\big(S | \calG_H(\bx_{\tiny \mbox{test}}) \big)$.
    \end{enumerate}
\end{proposition}

\paragraph{Implications for ICL-based models.}  While not strictly ruling out any benefit of trainable encoders, Proposition \ref{prop:labels_and_feature_importance} does mute confidence in their ability to robustly screen uninformative columns through the use of local neighborhood labels as additional inputs.  This plausibly suggests that, at least for RDBs with large numbers of feature columns (many of which are likely uninformative), trainable encoders may be unnecessary, especially given that parameter-free encoders are already structured to naturally handle column-wise feature screening \citep{JUICE_paper_2026}.  Therefore, consistent with our conclusions from Appendix \ref{app:labels_as_features}, TFMs with an appropriate $g_{\tiny \mbox{pf}}$ need not fall behind, even when considering encoder label usage for assessing task-specific feature importance.  

We also remark in closing that the RDB FM architecture described in \citet{hudovernik2026kumorfm} involves first converting each RDB table $\bT^k$ into revised embeddings $\bU^k$ dependent on task-specific labels, prior to formation of $\calG_H$ and subsequent presentation to $g_\theta$ (this conversion is accomplished with an additional module akin to a traditional TFM applied on single tables).  While certainly an innovative idea, and one that can be applied within purely TFM-based pipelines like RDBLearn as well, the extent to which uninformative columns can be resolved at this early stage may still have limitations (e.g., the counter-example used to prove Proposition \ref{prop:labels_and_feature_importance} would disallow filtering uninformative columns using the signal within a given $\bU^k$ alone).  However, further treatment of these possibilities lies beyond the scope of this work.

\section{Additional ICL Sampling Considerations}

\subsection{Evaluating Sample Complexity} \label{app:ICL_sampling_stuff}

We remark that \citet{ranjan2026rt,ranjanlarge} emphasize the impressive sample efficiency of RT-J relative to TFM-based RDBLearn, ostensibly requiring many fewer labeled ICL samples to achieve competitive performance.  But there are two key factors that merit consideration when interpreting the empirical comparisons upon which this claim is based:

\paragraph{Different notions of sample complexity.} ICL samples in  \citet{ranjan2026rt,ranjanlarge} are only counted per test instance, as opposed to counting the \textit{total} number of unique samples required across an entire test set, i.e., the union of unique samples extracted over all test instances associated with a given task.  This distinction is critical, since RDBLearn is designed to adopt a \textit{fixed} ICL batch that is shared over all test instances, while RT-J requires a \textit{new} batch for each separate inference point. Obviously then, the cardinality of the fixed RDBLearn set will generally be larger than what RT-J adopts for just a single test instance.  But if we compute the average number of \textit{unique} ICL samples needed per test instance, RDBLearn will often operate using far fewer depending on the RDB task (see empirical test results in Table \ref{tab:label_usage_comparison} below).  Hence RDBLearn can potentially be  more sample efficient relative to RT-J per this alternative metric as the number of test points grows larger.\footnote{Even with just a single test point, there is an important caveat worth mentioning.  While RT-J can operate successfully with relatively few labeled samples for any given test point, these samples must be reachable via random walks in the neighborhood of the associated target row.  Consequently, if the target row is to be selected without any prior restrictions, we must implicitly assume that our RDB has sufficient labeled rows around any candidate target, which then implies that we actually have access to a much larger overall collection of labels.}  This distinction impacts not only the required number of labeled observations, but also execution costs, since RDBLearn need only extract ICL samples and construct a fixed TFM KV cache once per task, regardless of how many test points exist.

\vspace*{-0.1cm}
\paragraph{Model alignment.} The actual per-instance empirical ICL sample complexity results reported in  \citet{ranjan2026rt,ranjanlarge} are not directly based on an  RDBLearn specification from either \citet{JUICE_paper_2026} or \citet{xu2026parameter}.  Instead, the RDBLearn pipeline is restricted to the TFM base model TabICL-v2 and paired with much smaller per-instance ICL sample batches originally prepared for RT-J (via random-walks combined with BFS local searches).   Notably, RT-J pretraining is explicitly aligned with this particular ICL-sampling regime, while RDBLearn is nominally designed for larger fixed ICL sets (per motivations from Section \ref{sec:locality}).  In Table \ref{tab:label_usage_comparison} below we empirically show that this distinction is important, as RDBLearn with uniform sampling outperforms random walk sampling, applied to either RT-J or RDBLearn, even while relying on \textit{fewer} labeled target instances.   Even so, our central point here is not to discount the prior comparisons between RT-J and RDBLearn from \citet{ranjan2026rt,ranjanlarge}, which certainly convey a useful complementary perspective, particularly if the goal is quantifying minimal sample complexity costs for a single test point executing from scratch.  Rather, we are simply advocating that additional factors provide important balance, especially when weighing diverse tasks involving a multiplicity of test points.

\vspace*{-0.1cm}
\paragraph{Empirical comparisons.}  Prior work has directly compared RT-J with random-walk (RW) versus uniform sampling; see Figure 6 in \citet{ranjan2026rt}.  Not surprisingly, from these results we observe that RW is superior since the RT-J model is explicitly pre-trained to exploit this format.  But there has not as of yet been any analogous experiment involving RDBLearn with RW versus uniform sampling; importantly, this requires a shared base model for meaningful head-to-head comparison.  To this end, we follow conditions from \citet{ranjan2026rt} and run RDBLearn with only TabICL-v2 as the base TFM.  We also execute across all RelBench-v1 tasks so that aggregated results are directly comparable with \citet{ranjan2026rt}.  For the uniformly sampled RDBLearn model, we follow default settings to form a fixed/shared context whereby the number of labels per task equals $\min \left( 10000, \mbox{train\_set\_size} \right)$.  Meanwhile, for all RW models we count the number of unique labels generated by the RT-J sampler (at two context sizes, 256 and 8K cells) aggregated over each test set instance (any duplicates are not counted).  Results are displayed in Table \ref{tab:label_usage_comparison}, from which we observe that RDBLearn, along with uniform sampling, is able to achieve the highest classification performance.   Moreover, this is accomplished even while relying on a small fraction of the total per-task labels that the RW models have access to.

\begin{table}[H]
\small
\centering
\setlength{\tabcolsep}{5.5pt}
\renewcommand{\arraystretch}{1.15}
\caption{Sample complexity comparison averaged across all 12 RelBench-v1 classification tasks (bold = best).  Approximate numbers ($\sim$) were estimated from Figure 3 in \citet{ranjan2026rt}.}
\label{tab:label_usage_comparison}
\begin{tabular}{l|cc|cc}
\toprule
\textbf{Method} & \shortstack{ \textbf{Sampling} \\ \textbf{Type} } & \shortstack{ \textbf{RW Context} \\ \textbf{Size/Cells} } & \shortstack{\textbf{Average} \\ \textbf{AUROC} $\left(\uparrow \right)$ } & \shortstack{\textbf{Avg Unique Labels}\\ \textbf{Used per Task} $\left(\downarrow \right)$}  \\
\midrule
RT-J          & RW & 256 & $\sim$ 69 & 22197  \\
RT-J          & RW & 8K & $\sim$ 73 & 83020  \\
RT-J (w/ ensembling)              & RW & 8K & 74.45 & 95079 \\
\midrule
RDBLearn (w/ TabICLv2)         & RW & 256 & $\sim$ 65 & 22197  \\
RDBLearn (w/ TabICLv2)         & RW & 8K & $\sim$ 72 & 83020  \\
\midrule
RDBLearn (w/ TabICLv2)         & Uniform & N/A & \textbf{75.0} & \textbf{7816}  \\
\bottomrule
\end{tabular}
\end{table}

\vspace*{-0.1cm}
\subsection{Degeneracy Example from Random Walk Sampling} \label{app:random_walk_example}
\vspace*{-0.1cm}

Assume data drawn according to Definition \ref{def:RDB_gen_process}, with original feature table $\bT^1 = \bX$ and corresponding labels $\by$, as well as induced auxiliary table $\bT^2 \in \calC^B$ composed of a single PK column.  The latter is assumed to correspond with a categorical column in $\bT^1$ that we treat as an FK column pointing to the PKs in $\bT^2$.  Granting this simplified setup, we observe that any random walk originating at an arbitrary row $i$ of $\bT^1$ will at most reach $|\calR_i |$ rows of $\bT^1$.  Here $\calR_i$ denotes the set of rows within $\bT^1$ that share the same FK value as row $i$.  There is no mechanism for longer random walks to eventually extend across the entire RDB; each must stall within $\calR_i$ because ultimately the underlying graph $\calG$ extracted from the RDB is  not fully connected.  

The graph underlying the experiments shown in Figure \ref{fig:interaction_full} exhibits an analogous phenomenon.  The only difference is that the FK column switches from (implicitly) $\bT^1$ to (explicitly) $\bT^2$ in this example, while the corresponding $\calG$ remains disconnected.  This might help explain why the RT-J model does not happen to work as well within the particular context of the Figure \ref{fig:interaction_full} experiments (note that RDBLearn would also not perform well here if forced to adopt the same RW sampling procedure).  That being said, if the RT-J context budget is not filled by random-walk accessible rows, a fallback involving hybrid uniform sampling akin to RDBLearn could represent a potential workaround.

\vspace*{-0.2cm}
\section{Language Model Agent Usage}  \label{app:LLM_usage}
\vspace*{-0.2cm}

A growing body of evidence suggests that LLM agents can play a useful role in tabular model development \citep{han2024large,huang2026relagent,kim2026refuge,zhang2024elf}.  That being said, their usage in the specific context of RDB predictive modeling involves notable caveats from a benchmarking perspective.  One in particular relates to potential leakage issues that may obscure the degree to which we can plausibly extrapolate benchmarking results to expected performance within new siloed industry settings.  This concern stems from the fact that existing public benchmarks are scarce, largely because of privacy concerns with real-world data.  And yet predictive solutions addressing the few RDB datasets that do exist, often involving feature engineering and human expertise, are readily visible to frontier LLM agents during their pre-training process or during inference-time retrieval.  Hence agent-based solutions may simply lean on problem-specific prior human knowledge to a degree that may not be possible on less-familiar or esoteric tasks within private RDB repositories.   Of course this is not meant to discourage LLM-based RDB frameworks, and certainly TFMs equipped with $g_{\tiny \mbox{pf}}$ (e.g., RDBLearn, TabPFN-Rel-OSS, etc.) are a natural fit for agentic enhancements, the topic of Section \ref{sec:agents}.  We only advocate that some caution is warranted when interpreting results, especially when extrapolating to new domains.  While similar concerns have certainly been raised with other data modalities, we would argue that the situation is more acute in RDB regimes with only limited, relatively well-known benchmarking sources that are susceptible to memorized solutions.

\section{Experiment Details and Full Results Behind Figure \ref{fig:big_comparison} } \label{app:experiment_details}

\subsection{Description of TFM-based Approaches}

 For TFM-based pipelines we adopt \textbf{RDBLearn} (version 1.1 from \citet{xu2026parameter}) and \textbf{TabPFN-Rel} (version from \citet{hayler2026advancing}); we use \textbf{TabPFN-Rel-OSS} to refer to the open-source release of TabPFN-Rel.\footnote{Improved configurations of both RDBLearn and TabPFN-Rel are possible, but lie outside of our scope.}  Since our goal is to examine TFM-based approaches on a wider task set than original papers considered, we ran both RDBLearn and TabPFN-Rel-OSS out of the box ourselves from the respective public repos using default configurations.\footnote{An exception is TabPFN-Rel-OSS on RelBench-v1 tasks, whereby existing results were pulled directly from \citet{hayler2026advancing} rather than running ourselves.}  Both methods were run on a single Nvidia RTX 5090 (32G) GPU. For RDBLearn, we encountered OOM errors in some tasks when the TFM was LimiX or TabFM. These sweep cases were simply skipped, so RDBlearn just selects from the remaining healthy cases. Meanwhile, we found that TabPFN-Rel-OSS did not execute on the multi-class SALT benchmark in our environment; hence this is our only exclusion in terms of reported results.\footnote{TabPFN-Rel-OSS presently does not support class cardinalities that exceed the base TFM capability, and SALT tasks happen to exceed this limit for now.} 
 
 In terms of architecture, RDBLearn and TabPFN-Rel-OSS are conceptually the same, with a shared $g_{\tiny \mbox{pf}}$ implementation, but differing primarily in which tabular base models are chosen and details of how ICL sample sets are constructed.  For example, TabPFN-Rel-OSS  includes validation samples in the final chosen ICL set as in \citet{hudovernik2026kumorfm}, and also oversamples ICL instances with more recent past timestamps; the latter is consistent with benefits of recency featurization bias explored by \citet{huang2026relagent}.  Neither RDBLearn nor TabPFN-Rel-OSS include text-column handling (text columns are simply ignored), unlike the closed-source TabPFN-Rel version (see below) which includes a proprietary module for this purpose.
 
We also remark that new crops of advanced single-table FMs are constantly being released; recent examples that appeared as we were finalizing this draft include TabPFN-v3.5 \citep{jager2026tabpfn35}, EXAONE \citep{eo2026exaone},  LimiX-2 \citep{zhang2026limix}, Mitra-v2 \citep{tao2026mitra}, and Xiaomi-TabLDM \citep{team2026xiaomi}.  These approaches can readily be combined with an encoder $g_{\tiny \mbox{pf}}$ to further boost performance relative to alternative RDB FM formulations.  However, we do not rerun all experiments using these options, since our central goal herein is not to surface the best possible TFM configuration within RDB operational settings.  The latter represents a moving target, and remains somewhat orthogonal to our primary objective, which the results from Figure \ref{fig:big_comparison} already adequately achieve.

\subsection{Description of Other Baseline Models}

To compare against the TFM-based approaches from above, we present evaluations across a wide range of alternative RDB predictive architectures, including both foundation models and more traditional supervised learning approaches.  We emphasize though that results are not presently available for all models on all datasets.  For example, it is frequent practice to only report results on RelBench-v1 datasets and tasks, or to only report results on some tasks within a given benchmark.

\paragraph{Closed-Source Foundation Models.} We include the original industry model \textbf{KumoRFM-1} \citep{fey2025kumorfm} as well as the more recent version \textbf{KumoRFM-2} \citep{hudovernik2026kumorfm} that incorporates task-specific labels within the encoder and constituent table embeddings.\footnote{KumoRFM-2 has generally been treated as a closed-source model and we maintain this designation here given that all KumoRFM-2 results in Figure \ref{fig:big_comparison} are based on private API calls.  Moreover, the KumoRFM-2 pre-training regime and extent of real-world RDB exposure remain undisclosed.}   We also compare with \textbf{OpenRFM} \citep{chen2026openrfm}; while this model is presumably intended to be open-sourced at some point, the full code is not available at the time of this writing.  For all of these closed-source approaches we rely on posted performance results from original authors for evaluation purposes, with one exception.  We follow \citet{grinsztajn2026tabpfn3}, who have  independently reevaluated KumoRFM-2 performance via API calls for certain datasets using author-provided benchmarking scripts.

\paragraph{Open-Source Foundation Model Baselines.}  Open-source baselines  include the schema-agnostic \textbf{Griffin} \citep{wanggriffin}, \textbf{RelLLM} \citep{wu-etal-2025-large}, and \textbf{RT} (relational transformer) \citep{ranjan2026relational} models, as well as an updated RT version enhanced with \textbf{PluRel} synthetic pretraining data \citep{kothapalli2026plurel}, denoted \textbf{RT-P}. 
On the open-source ICL-based side we include results with \textbf{RT-J}, also based on the RT architecture but with modifications such as random-walk sampling to collect ICL sets (as discussed in Section \ref{sec:random_walk_sampling} of the main text) and pretraining with a large collection of real-world RDBs called \textbf{The Join} \citep{ranjan2026rt}; see also Appendix \ref{app:ICL_sampling_stuff} for additional perspective on RT-J sampling considerations and prior comparisons with RDBLearn.  Furthermore, we compare results from \textbf{RDB-PFN} 
\citep{wang2026relational}, with synthetic RDB pre-training, and \textbf{RandGNN} \citep{JUICE_paper_2026}, which combines a random-weight GNN encoder with a TFM-based model.  RandGNN represents a form of ablation w.r.t.~RDBLearn and TabPFN-Rel-OSS, in that it explicitly mixes cross-column information in a manner at odds with the theoretical considerations upon which these models are based.  And finally, for pure language model baselines, we evaluate \textbf{LLM-A} \citep{team2025gemma} (as tested in \citet{ranjan2026relational}) and \textbf{LLM-B} \citep{wydmuch2024tackling}; both approaches are predicated on serialized RDB neighborhood representations and/or ICL samples.  Although the underlying LLMs are not open-source, we still treat these approaches as open-source given that the prompts used to reproduce results are publicly available and no hidden fine-tuning steps are involved.

\paragraph{Task-Specific Supervised Models.}  While our emphasis is on RDB foundation models, it is still useful to contextualize FM performance results w.r.t.~methods benefitting from per-task supervised training.  On this front, largely following prior work, we include the more classical standards \textbf{LightGBM}, \textbf{XGBoost}, and \textbf{GraphSage} as evaluated in \citet{hudovernik2026kumorfm}.  Meanwhile, for recent SOTA supervised architectures we select \textbf{RelGNN} \citep{chen2025relgnn}, \textbf{RelGT} \citep{dwivedi2025relational}, \textbf{RelAgent} \citep{huang2026relagent}, and \textbf{RT-J+fine-tuning}; the latter introduces task-specific fine-tuning of the pre-trained RT-J architecture mentioned previously to achieve SOTA performance on RelArena-$\alpha$ \citep{hayler2026advancing}.

\subsection{RDB Benchmarks}  

In producing Figure \ref{fig:big_comparison}, we evaluate RDB predictive modeling approaches over a diverse suite of benchmarks, including regression and binary classification tasks from both \textbf{RelBench-v1} \citep{robinson2024relbench} and \textbf{RelBench-v2}  \citep{gu2026relbench}.  Note that RelBench-v1 datasets and tasks are also equivalent to the recently-posted \textbf{RelArena-}$\balpha$ benchmarking platform \citep{hayler2026advancing} patterned after TabArena \citep{erickson2026tabarena}.  The primary difference is that RelArena-$\alpha$ enforces more rigid/standardized submission protocols that some published work on RelBench-v1 may not satisfy.  That being said, RDBLearn and TabPFN-Rel-OSS already adhere to these guidelines, and it is only to the advantage of other models with which we compare if they do not.  This means our results in Figure \ref{fig:big_comparison} are actually conservative in showcasing the abilities of TFM-based approaches to RDB predictive modeling.

Beyond the above, we test over five \textbf{4DBInfer} \citep{wang20244dbinfer} binary classification datasets that have been used in several recent works \citep{huang2026relagent,hudovernik2026kumorfm}, as well as the multi-class predictive tasks from the \textbf{SALT} repository \citep{klein2024salt} as adopted by \citet{hudovernik2026kumorfm}.  Standard train/dev/test splits are used for all experiments. Note that for RelBench-v1, we omit only rel-f1 and rel-event datasets because of potential leakage issues mentioned in prior work, and we do not have results for all baseline models on these datasets.

\subsection{Extended Empirical Results} \label{app:extended_results}

In this section we provide the full set of experimental results upon which the final rankings shown in Figure \ref{fig:big_comparison} from the main text are based.  The TFM-based methods (along with their attendant $g_{\tiny \mbox{pf}}$) achieve the best overall performance among RDB FMs.  This follows by comparing numbers for RDBLearn, TabPFN-Rel-OSS, and TabPFN-Rel versus other RDB FMs in the tables below.  Surprisingly, these frameworks even exceed the results produced by many per-task supervised models.

\begin{table}[H]
\small
\centering
\setlength{\tabcolsep}{5.5pt}
\renewcommand{\arraystretch}{1.15}
\caption{Results on 4DBInfer binary classification tasks (AUC$\uparrow$). Bold = best per column, underline = second best.}
\label{tab:4dbinfer}
\begin{tabular}{l|ccccc|cc}
\toprule
\textbf{Method} & \shortstack{AB\\churn} & \shortstack{OB\\ctr} & \shortstack{RR\\cvr} & \shortstack{SE\\upvote} & \shortstack{SE\\churn} & \shortstack{\textbf{Avg AUROC}\\$\uparrow$} & \shortstack{\textbf{Avg Rank}\\$\downarrow$} \\
\midrule
\multicolumn{8}{l}{\textbf{Supervised}}\\
XGBoost          & 50.00 & 50.00 & 50.00 & 49.68 & 50.84 & 50.10 & 9.6 \\
GNN              & 75.71 & 62.39 & 84.70 & 88.61 & 85.58 & 79.40 & 4.6 \\
RelAgent         & \textbf{79.44} & 62.63 & 86.35 & \textbf{89.45} & \underline{89.03} & 81.38 & \textbf{2.0} \\
\midrule
\multicolumn{8}{l}{\textbf{Foundation model}}\\
\multicolumn{8}{l}{\quad\textit{Closed-Source}}\\
KumoRFM-1        & 75.59 & 61.11 & 84.36 & \underline{88.92} & 87.87 & 79.57 & 5.1 \\
KumoRFM-2        & 77.06 & 61.97 & 84.36 & 88.50 & 87.92 & 79.96 & 4.9 \\
RDB-PFN          & 73.38 & 52.59 & 78.12 & 85.60 & 85.50 & 75.04 & 7.2 \\
\addlinespace[2pt]
\multicolumn{8}{l}{\quad\textit{Open-source}}\\
Griffin                  & 53.32 & 52.22 & 39.92 & 71.66 & 72.18 & 57.86 & 8.8 \\
RandGNN                  & 77.42 & 49.29 & 66.21 & 61.06 & 77.41 & 66.28 & 7.6 \\
TabPFN-Rel-OSS           & 77.21 & \underline{63.68} & \underline{88.92} & 88.59 & \textbf{89.14} & \underline{81.51} & \underline{2.6} \\
RDBLearn                 & \underline{78.71} & \textbf{63.76} & \textbf{89.06} & 87.86 & 88.69 & \textbf{81.62} & \underline{2.6} \\
\bottomrule
\end{tabular}
\end{table}

\begin{table}[h]
\centering
\small
\setlength{\tabcolsep}{5pt}
\renewcommand{\arraystretch}{1.15}
\caption{Results on the SALT benchmark (MRR$\uparrow$). Bold = best per column; underline = second best.}\label{tab:salt}
\resizebox{\columnwidth}{!}{%
\begin{tabular}{l|ccc|ccccc|cc}
\toprule
& \multicolumn{3}{c}{salt-item} & \multicolumn{5}{c}{salt-sales} & \textbf{Avg MRR} & \textbf{Avg Rank} \\
\cmidrule(lr){2-4} \cmidrule(lr){5-9}
\textbf{Method} & Plant & \shortstack{Shipping\\Point} & Incoterm
       & Incoterm & Office & Group & \shortstack{Payment\\Terms} & \shortstack{Shipping\\Condition}
       & $\uparrow$ & $\downarrow$ \\
\midrule
\multicolumn{11}{l}{\textbf{Supervised}}\\
GraphSAGE  & \underline{0.99} & \underline{0.98} & 0.69
           & 0.62 & \textbf{1} & 0.16 & 0.37 & 0.57
           & 0.67 & 3.88 \\
LightGBM  & 0.61 & 0.28 & 0.73
           & 0.73 & \underline{0.99} & 0.02 & 0.10 & 0.51
           & 0.50 & 4.75 \\
\midrule
\multicolumn{11}{l}{\textbf{Foundation model}}\\
\multicolumn{11}{l}{\quad\textit{Closed-Source}}\\
KumoRFM-1  & \underline{0.99} & \textbf{0.99} & \underline{0.79}
           & \underline{0.81} & \textbf{1} & 0.38 & 0.38 & 0.78
           & 0.77 & \underline{2.56} \\
KumoRFM-2  & \underline{0.99} & \underline{0.98} & 0.78
           & \underline{0.81} & \textbf{1} & \underline{0.61} & \underline{0.61} & \underline{0.79}
           & \underline{0.82} & \underline{2.56} \\
\addlinespace[2pt]
\multicolumn{11}{l}{\quad\textit{Open-source}}\\
RDBLearn  & \textbf{1} & \textbf{0.99} & \textbf{0.82}
           & \textbf{0.91} & \textbf{1} & \textbf{0.69} & \textbf{0.79} & \textbf{0.84}
           & \textbf{0.88} & \textbf{1.25} \\
\bottomrule
\end{tabular}%
}
\end{table}

\begin{table}[h]
\centering
\caption{Results on Relbench v2 binary classification tasks (AUROC$\uparrow$). Bold = best per column; underline = second best.}\label{tab:relbenchv2_classfication}
\setlength{\tabcolsep}{5.5pt}
\renewcommand{\arraystretch}{1.15}
\resizebox{\columnwidth}{!}{%
\begin{tabular}{l|ccccc|cc}
\toprule
& mimic & \multicolumn{3}{c}{ratebeer} & arxiv & \textbf{Avg AUROC} & \textbf{Avg Rank} \\
\cmidrule(lr){2-2} \cmidrule(lr){3-5} \cmidrule(lr){6-6}
\textbf{Method} & stay & beer-churn & user-churn & dormant & citation & $\uparrow$ & $\downarrow$ \\
\midrule
\multicolumn{8}{l}{\textbf{Supervised}}\\
LightGBM             & 51.81 & 76.21 & 83.92 & 75.79 & 71.21 & 71.79 & 5.80 \\
GraphSAGE            & 55.01 & 78.67 & 94.27 & 80.51 & 82.50 & 78.19 & 3.60 \\
Rel-Agent$^{\dagger}$ & -- & \underline{84.70} & 97.43 & \textbf{83.33} & \underline{82.61} & -- & -- \\
\midrule
\multicolumn{8}{l}{\textbf{Foundation model}}\\
\multicolumn{8}{l}{\quad\textit{Closed-Source}}\\
KumoRFM-1            & \textbf{56.33} & 75.06 & 91.38 & 77.10 & 80.62 & 76.10 & 4.40 \\
KumoRFM-2            & 55.28 & 83.84 & 97.43 & 80.65 & 81.71 & \underline{79.78} & 3.00 \\
OpenRFM (TabICL)$^{\dagger}$ & --        & 79.60             & --                & 77.50             & 81.00             & --                & --   \\
OpenRFM (TabPFN)$^{\dagger}$ & --        & 80.90             & --                & 79.10             & 81.00             & --                & --   \\
\addlinespace[2pt]
\multicolumn{8}{l}{\quad\textit{Open-source}}\\
TabPFN-Rel-OSS       & 53.10 & 80.15 & \underline{98.28} & \underline{82.32} & \textbf{82.69} & 79.31 & \underline{2.40} \\
RDBLearn             & \underline{55.36} & \textbf{85.15} & \textbf{98.48} & 81.85 & 82.37 & \textbf{80.64} & \textbf{1.80} \\
\bottomrule
\end{tabular}%
}
\vspace{2pt}
\begin{minipage}{\columnwidth}
\raggedright\footnotesize
$^{\dagger}$\,RelAgent and OpenRFM are excluded when calculating average AUROC and rank because results are missing in some cases (RelAgent: no mimic-stay; OpenRFM: no mimic-stay or ratebeer user-churn). Note also that public code is not available to produce results for the missing tasks at the time of this writing.
\end{minipage}
\end{table}

\begin{table}[h]
\centering
\caption{Results on Relbench v1 binary classification tasks (AUROC$\uparrow$). Bold = best per column, underline = second best.}
\setlength{\tabcolsep}{5.5pt}
\renewcommand{\arraystretch}{1.15}
\resizebox{\columnwidth}{!}{%
\begin{tabular}{l|cccccccc|cc}
\toprule
& \multicolumn{2}{c}{avito} & trial & \multicolumn{2}{c}{amazon} & \multicolumn{2}{c}{stack} & hm & \textbf{Avg AUROC} & \textbf{Avg Rank} \\
\cmidrule(lr){2-3} \cmidrule(lr){4-4} \cmidrule(lr){5-6} \cmidrule(lr){7-8} \cmidrule(lr){9-9}
\textbf{Method} & click & visit & out & user & item & eng & badge & churn & $\uparrow$ & $\downarrow$ \\
\midrule
\multicolumn{11}{l}{\textbf{Supervised}}\\
LightGBM                & 53.60             & 53.05             & 70.09             & 52.22             & 62.54             & 63.39             & 63.43             & 55.21             & 59.19             & 18.88 \\
GraphSAGE               & 65.90             & 66.20             & 68.60             & 70.42             & 82.81             & 90.59             & 88.86             & 69.88             & 75.41             & 6.38 \\
RelGT                   & \underline{68.30} & 66.78             & 68.61             & 70.39             & 82.55             & 90.53             & 86.32             & 69.27             & 75.34             & 6.81 \\
RelGNN                  & 68.23             & 66.18             & 71.24             & \underline{70.99} & 82.64             & \textbf{90.75}    & \underline{88.98} & \underline{70.93} & 76.24             & 4.12 \\
RelAgent                & \textbf{68.36}    & \underline{67.79} & 71.86             & 70.78             & \underline{82.84} & 90.41             & 88.42             & \textbf{71.07}    & \underline{76.44} & \textbf{3.50} \\
RT-J+fine-tuning        & 66.81             & 66.89             & 69.57             & \textbf{71.46}    & \textbf{83.16}    & \underline{90.65} & \textbf{89.21}    & 70.49             & 76.03             & 4.00 \\
\midrule
\multicolumn{11}{l}{\textbf{Foundation model}}\\
\multicolumn{11}{l}{\quad\textit{Closed-Source}}\\
KumoRFM-1               & 64.85             & 64.11             & 70.79             & 67.29             & 79.93             & 87.09             & 80.00             & 67.71             & 72.72             & 11.56 \\
KumoRFM-2 (orig.)*      & 67.42             & \textbf{69.41}    & 72.03             & 69.10             & 82.17             & 89.40             & 87.15             & 69.27             & 75.74             & 6.62 \\
KumoRFM-2 (repro.)*     & 67.42             & \textbf{69.41}    & 72.03             & 67.71             & 80.18             & 88.69             & 85.40             & 67.81             & 74.83             & 7.69 \\
TabPFN-Rel              & 67.52             & 66.80             & \textbf{76.47}    & 70.86             & 82.80             & 90.60             & 88.04             & 70.52             & \textbf{76.70}    & \underline{3.88} \\
OpenRFM (TabICL)$^{\dagger}$ & 62.30 & 64.60 & 60.80 & -- & -- & 88.50 & 86.30 & 68.60 & -- & -- \\
OpenRFM (TabPFN)$^{\dagger}$ & 63.90 & 64.60 & 63.40 & -- & -- & 88.40 & 86.40 & 68.90 & -- & -- \\
\addlinespace[2pt]
\multicolumn{11}{l}{\quad\textit{Open-source}}\\
LLM-A                   & 59.80             & 62.70             & 57.40             & 58.10             & 62.10             & 78.00             & 80.00             & 59.80             & 64.74             & 17.69 \\
LLM-B                   & 61.32             & 60.28             & 55.72             & 60.56             & 71.96             & 81.01             & 71.13             & 64.34             & 65.79             & 17.12 \\
RelLLM                  & 62.28             & 56.17             & 59.02             & 60.07             & 64.10             & 69.46             & 62.12             & 55.95             & 61.15             & 18.25 \\
Griffin                 & 45.90             & 60.70             & 51.00             & 62.30             & 69.00             & 77.50             & 73.50             & 60.20             & 62.51             & 18.62 \\
RT                      & 59.50             & 61.80             & 51.80             & 64.00             & 70.90             & 75.70             & 80.10             & 62.80             & 65.83             & 17.19 \\
RT-J                    & 51.39             & 64.09             & 64.43             & 67.20             & 78.61             & 86.76             & 81.11             & 65.66             & 69.91             & 14.25 \\
RT-P                    & 47.90             & 63.40             & 51.80             & 65.00             & 72.50             & 86.20             & 82.00             & 66.00             & 66.85             & 15.69 \\
RDBPFN                  & 64.46             & 65.82             & 64.50             & 65.92             & 79.59             & 88.04             & 84.52             & 67.21             & 72.51             & 11.75 \\
RandGNN                 & 61.58             & 64.15             & 55.98             & 65.87             & 77.73             & 85.14             & 83.14             & 66.54             & 70.02             & 13.75 \\
TabPFN-Rel-OSS          & 61.45             & 66.88             & \underline{73.06} & 70.24             & 82.79             & 90.58             & 86.35             & 70.57             & 75.24             & 6.00 \\
RDBLearn                & 67.55             & 66.71             & 72.71             & 69.33             & 82.25             & 89.92             & 85.23             & 68.27             & 75.25             & 7.25 \\
\bottomrule
\end{tabular}%
}
\vspace{2pt}
\begin{minipage}{\columnwidth}
\raggedright\footnotesize
$^{*}$\,Results are drawn from both the original KumoRFM-2 paper \citep{hudovernik2026kumorfm}, as well as an independent reproduction from \citet{grinsztajn2026tabpfn3} using KumoRFM-2 API calls for some datasets.  (The latter relies on official benchmarking scripts for setting the various KumoRFM-2 hyperparameters that may vary from task to task.  For datasets with no available scripts, avito and trial, the reproduction value is set to the original.)
$^{\dagger}$\,OpenRFM is excluded when calculating average AUROC and rank because results are not available for all datasets, and public code is not available to produce results for the missing tasks.
\end{minipage}
\label{tab:relbenchv1_classification}
\end{table}

\begin{table}[h]
\small
\centering
\caption{Results on Relbench-v1 regression tasks (MAE$\downarrow$). Bold = best per column; underline = second best.}\label{tab:relbenchv1_regression}
\setlength{\tabcolsep}{5.5pt}
\renewcommand{\arraystretch}{1.15}
\resizebox{\columnwidth}{!}{%
\begin{tabular}{l|ccccccc|cc}
\toprule
& avito & \multicolumn{2}{c}{trial} & \multicolumn{2}{c}{amazon} & stack & hm & \textbf{Geo Mean} & \textbf{Avg Rank} \\
\cmidrule(lr){2-2} \cmidrule(lr){3-4} \cmidrule(lr){5-6} \cmidrule(lr){7-7} \cmidrule(lr){8-8}
\textbf{Method} & ctr & adverse & success & user & item & votes & sales & $\downarrow$ & $\downarrow$ \\
\midrule
\multicolumn{10}{l}{\textbf{Base}}\\
Global Median           & 0.043             & 57.533            & 0.462             & 16.783            & 64.234            & 0.068             & 0.076             & 1.303             & 14.57 \\
Entity Median           & 0.046             & 57.930            & 0.441             & 17.423            & 66.436            & 0.069             & 0.078             & 1.329             & 15.57 \\
\midrule
\multicolumn{10}{l}{\textbf{Supervised}}\\
LightGBM                & 0.041             & 44.011            & 0.425             & 16.783            & 60.569            & 0.068             & 0.076             & 1.220             & 12.50 \\
GraphSAGE               & 0.041             & 44.473            & 0.400             & 14.313            & 50.053            & \underline{0.065} & 0.056             & 1.096             & 8.93 \\
RelGT                   & 0.035             & 43.992            & \underline{0.326} & 14.267            & 48.922            & \underline{0.065} & 0.054             & 1.030             & 6.57 \\
RelGNN                  & 0.037             & 44.461            & \textbf{0.301}    & 14.230            & 48.767            & \underline{0.065} & 0.054             & 1.027             & 6.64 \\
RelAgent                & 0.033             & \underline{37.194} & 0.386             & \underline{13.949} & \textbf{41.765}   & \textbf{0.064}    & \underline{0.035} & \textbf{0.934}    & \textbf{2.57} \\
RT-J+fine-tuning        & 0.0345            & \textbf{34.9}     & 0.364             & 13.97             & \underline{42.49} & 0.0653            & 0.0375            & \underline{0.937} & \underline{3.86} \\
\midrule
\multicolumn{10}{l}{\textbf{Foundation Model}}\\
\multicolumn{10}{l}{\quad\textit{Closed-Source}}\\
KumoRFM-1               & 0.035             & 58.231            & 0.417             & 16.161            & 55.254            & \underline{0.065} & 0.040             & 1.102             & 9.50 \\
KumoRFM-2 (orig.)*      & 0.034             & 43.293            & 0.433             & \textbf{13.921}   & 46.992            & \textbf{0.064}    & \textbf{0.034}    & 0.986             & 4.86 \\
KumoRFM-2 (repro.)*     & 0.033             & 41.974            & 0.433             & 14.627            & 45.352            & \underline{0.065} & 0.043             & 1.015             & 6.21 \\
TabPFN-Rel              & \textbf{0.0311}   & 39.8              & 0.413             & 14.4              & 46.8              & 0.0679            & 0.0605            & 1.051             & 6.36 \\
\addlinespace[2pt]
\multicolumn{10}{l}{\quad\textit{Open-source}}\\
Griffin                 & 0.050             & 78.232            & 0.463             & 35.590            & 53.214            & 0.092             & 0.151             & 1.737             & 16.86 \\
RT                      & 0.058             & 73.999            & 0.455             & 18.802            & 57.996            & 0.110             & 0.089             & 1.543             & 16.71 \\
RT-J                    & 0.0402            & 55.377            & 0.4265            & 16.757            & 46.151            & 0.0752            & 0.0501            & 1.156             & 10.57 \\
RandGNN                 & 0.038             & 53.370            & 0.434             & 17.580            & 66.130            & 0.071             & 0.065             & 1.248             & 14.00 \\
TabPFN-Rel-OSS          & \underline{0.0314} & 42.6              & 0.386             & 14.4              & 47.8              & 0.068             & 0.0614            & 1.058             & 7.29 \\
RDBLearn                & 0.033             & 43.565            & 0.378             & 14.504            & 47.038            & 0.067             & 0.063             & 1.066             & 7.43 \\
\bottomrule
\end{tabular}%
}
\vspace{2pt}
\begin{minipage}{\columnwidth}
\raggedright\footnotesize
$^{*}$\,See comment from Table \ref{tab:relbenchv1_classification}.
\end{minipage}
\end{table}

\clearpage

\section{Setup Details and Full Empirical Results for Figure \ref{fig:single_table} } \label{app:continuum_experiment}

\vspace*{-0.1cm}
\paragraph{Setup.}  For our initial source of single tables, we select binary classification and regression tasks from BeyondArena~\citep{purucker2026beyond}, Fraud Dataset Benchmark (FDB) \citep{grover2022fraud}, and OpenML~\citep{vanschoren2014openml}. The origins of specific tasks are listed in Table~\ref{tab:single_table_results}. To better align with realistic RDBs, we consider tasks that have temporal splits, e.g., the ``Temporal'' category within BeyondArena. Large datasets (\textit{home\_credit\_default\_stability}) and datasets without categorical columns (\textit{maps\_router\_eta}) are excluded. To induce RDBs from these tables, we select at most $5$ categorical columns as FKs following steps 1--4 of the procedure described in Section~\ref{sec:continuum}.  As mentioned in the main text, we retain the original dataset labels as opposed to synthetically generating new ones.  This allows us to probe RDB FM behavior in regimes where relational structure exists, but is not strictly necessary for making accurate predictions. 

\vspace*{-0.1cm}
\paragraph{Extended results.}  In Table~\ref{tab:single_table_results} we present the full results that contributed to Figure \ref{fig:single_table}, including performance from TabPFN-v3 \citep{grinsztajn2026tabpfn3} applied only on the original single-table data as a reference point, RDBLearn~\citep{JUICE_paper_2026}, KumoRFM-2~\citep{hudovernik2026kumorfm}, and RT-J~\citep{ranjan2026rt}. Because BeyondArena has no validation set, and to standardize comparisons, we simply execute RDBLearn under the fixed setting of hop $H=2$ along with just a single TFM selected as TabPFN-v3.  For KumoRFM-2, we also use the same default hop value $H=2$ and leave the inference-time configuration unchanged. RT-J uses the default inference configuration as well. Regarding hardware, we ran each method on a single GPU. TabPFN-v3, RDBLearn, and RT-J were run on a single Nvidia RTX 5090 (32G). Although the dataset scale is the same, KumoRFM-2 requires more GPU memory. To fulfill this requirement, we ran KumoRFM-2 on a single Nvidia L20 (48G). However, there were still four tasks raising OOM errors, noted in Table~\ref{tab:single_table_results} and Figure \ref{fig:single_table}.

\begin{table}[h]
  \centering
  \caption{Full results corresponding to Figure~\ref{fig:single_table}, where we examine RDBs induced from single tables.
    Classification reports ROC-AUC ($\uparrow$);
    regression reports $R^{2}$ ($\uparrow$).
    Best result per row in \textbf{bold}; second best \underline{underlined}.
    Avg (excluding OOM) represents mean performance computed only over tasks/rows where every method reports a value;
    Avg (all) includes all tasks (KumoRFM-2 omitted due to OOM entries).}
  \label{tab:single_table_results}
  \resizebox{\columnwidth}{!}{%
  \begin{tabular}{ll|cccc}
    \toprule
    Task & Origin & TabPFN-v3 & RDBLearn & KumoRFM-2 & RT-J \\
    \midrule
    \multicolumn{6}{l}{\textit{Classification (ROC-AUC $\uparrow$)}} \\
    \midrule
    \texttt{acquire\_valued\_shoppers\_challenge} & BeyondArena & \textbf{0.6245} & \underline{0.6098} & OOM & 0.5797 \\
    \texttt{anes\_voting\_2026} & BeyondArena & \textbf{0.8620} & \underline{0.8260} & OOM & 0.5000 \\
    \texttt{bank\_marketing\_temporal\_month} & OpenML & \textbf{0.8856} & 0.8555 & \underline{0.8568} & 0.3762 \\
    \texttt{electricity\_temporal} & OpenML & \textbf{0.9542} & \underline{0.9098} & 0.8752 & 0.6176 \\
    \texttt{hotel\_booking\_demand} & BeyondArena & \textbf{0.8059} & \underline{0.7265} & 0.6645 & 0.5755 \\
    \texttt{kick} & BeyondArena & \textbf{0.8220} & \underline{0.7840} & 0.7611 & 0.5949 \\
    \texttt{kickstarter} & BeyondArena & \textbf{0.9101} & \underline{0.9061} & 0.8354 & 0.6860 \\
    \texttt{lending\_club} & BeyondArena & \textbf{0.6767} & \underline{0.6573} & 0.6311 & 0.5628 \\
    \texttt{sparknov\_temporal} & FDB & \textbf{0.9809} & \underline{0.9722} & 0.9079 & 0.6103 \\
    \midrule
    \textit{Avg (excluding OOM)} &  & \textit{\textbf{0.8622}} & \textit{\underline{0.8302}} & \textit{0.7903} & \textit{0.5748} \\
    \textit{Avg (all)} &  & \textit{\textbf{0.8358}} & \textit{\underline{0.8052}} & \textit{---} & \textit{0.5670} \\
    \midrule
    \multicolumn{6}{l}{\textit{Regression ($R^{2}$ $\uparrow$)}} \\
    \midrule
    \texttt{california\_house\_prices\_2020} & BeyondArena & \textbf{0.9351} & \underline{0.9345} & 0.9091 & 0.0026 \\
    \texttt{climate\_model\_weather\_forecasting} & BeyondArena & \textbf{0.8339} & \underline{0.8293} & 0.7879 & 0.7575 \\
    \texttt{coffee\_rating\_prediction} & BeyondArena & \underline{0.3390} & \textbf{0.3515} & 0.0928 & 0.0769 \\
    \texttt{cooking\_time} & BeyondArena & \textbf{0.4549} & \underline{0.4505} & 0.3617 & 0.3211 \\
    \texttt{delivery\_eta} & BeyondArena & \textbf{0.4375} & 0.3435 & \underline{0.4034} & 0.2797 \\
    \texttt{garments\_worker\_productivity} & BeyondArena & \underline{0.3686} & \textbf{0.5094} & 0.3011 & $-$0.1818 \\
    \texttt{mercedes\_benz\_greener\_manufacturing} & BeyondArena & \textbf{0.5519} & \underline{0.5446} & OOM & 0.0000 \\
    \texttt{rossmann\_store\_sales} & BeyondArena & \underline{0.8161} & \textbf{0.8244} & 0.5957 & 0.3062 \\
    \texttt{sberbank\_housing\_market\_forecasting} & BeyondArena & \textbf{0.6283} & \underline{0.5978} & OOM & 0.0149 \\
    \texttt{sf\_permit\_time} & BeyondArena & 0.0610 & \underline{0.1751} & 0.1669 & \textbf{0.2714} \\
    \midrule
    \textit{Avg (excluding OOM)} &  & \textit{\underline{0.5308}} & \textit{\textbf{0.5523}} & \textit{0.4523} & \textit{0.2292} \\
    \textit{Avg (all)} &  & \textit{\underline{0.5426}} & \textit{\textbf{0.5561}} & \textit{---} & \textit{0.1849} \\
    \bottomrule
  \end{tabular}%
  }
\end{table}

\clearpage
\section{Details of Interaction Experiments} \label{app:interaction_experiments} \label{app:interaction_details}

This section describes the detailed setup for experimenting with the interaction terms discussed in Sections \ref{sec:adequate_encoder} and \ref{sec:agents}, particularly as they relate to Figure \ref{fig:interaction_tests}.

\subsection{Setup Behind Figure \ref{fig:interaction_full} (left) }

\paragraph{Data generation.} \label{app:interaction_synthetic}
Each synthetic RDB contains two tables, a target/parent  $\bT^1$ and a child  $\bT^2$.  Table $\bT^1$ contains two columns (PK and a label) and 4000 rows (3200 for training and 800 for testing).  Each row of $\bT^1$ is linked to $l$ rows of $\bT^2$ via PK-FK pairs, where $\bT^2$ is composed of a single FK column as well as two signal columns $\{\ba, \bb \}$ mixed within a variable number of uninformative columns.  We set $l \sim \mbox{Uniform}\{2, \ldots, 20\}$. Signal columns are synthesized with Bernoulli(0.3) for binary variants or $\mathcal{N}(0,1)$ for Gaussian ones, alongside $d_2 - 2$ additional columns drawn IID from the same respective family, for total widths $d_2 \in \{8, 16, 32, 64, 96\}$ for $\bT^2$ instantiations.

\paragraph{Label synthesis.} We incorporate 4 labeling functions to cover both linear and nonlinear relationships. For each row in the task table where PK$=i$, let $ \calR_i=\{ j \in \bT^2 ~\text{where FK}_j=i \}$. Labels are generated by aggregating the row-level interaction with different the aggregation functions:
\begin{itemize}
    \item \texttt{mean}: ~ $y_i=\text{mean}_{j\in\calR_i} a_j b_j$ ;
    \item \texttt{thresholded count}: ~ $y_i = \text{count}_{j\in \calR_i}(a_jb_j>1)$ ;
    \item \texttt{parity}: ~ $y_i=\bigoplus_{j\in \calR_i} (a_j\land b_j$) ;
     \item \texttt{prod\_agg}: ~ $y_i=\mbox{mean}_{\calR_i}(\ba) \cdot \mbox{mean}_{\calR_i}(\bb)$.
\end{itemize}
We remark that \texttt{prod\_agg} is favorable to RDBLearn since $g_{\tiny \mbox{pf}}$ already computes this form of interaction between aggregations, as mentioned in Section \ref{sec:adequate_encoder}.  Meanwhile, the other aggregation choices are not favorable to RDBLearn since they are not computed by any single aggregation (although combinations thereof may approximate them).  For different labeling rules, we experimented with different data column distributions as follows:
\begin{itemize}
    \item \texttt{parity}, \texttt{mean}, and \texttt{prod\_agg} with binary columns;
    \item \texttt{count}$(a b > 1)$ with Gaussian columns.
\end{itemize}
Together these selections give us the four tasks listed in Figure \ref{fig:interaction_full} (\textit{left}).

\paragraph{Experiment models.} We compare TFM-based RDBLearn against the top non-TFM-based RDB FMs, KumoRFM-2 \citep{hudovernik2026kumorfm} and RT-J \citep{ranjan2026rt}, all in-context with no task-specific training or fine-tuning. RDBLearn simply uses $H=2$ to match the KumoRFM-2 default setting, run from a local deployment of the published model.  RT-J uses the public zero-shot checkpoint for FM operational mode (i.e., no fine-tuning).  For both KumoRFM-2 and RT-J we first validated our implementation pipeline by reproducing performance numbers from the respective published papers.

\subsection{Quasi-Real World RDB Used in Figure \ref{fig:interaction_full} (right)}

We adopt the rel-trial RDB \citep{robinson2024relbench} for the quasi-real world interaction experiment. We do not modify the actual schema; instead, for each trial we pick two numerical or categorical columns $\{\ba, \bb\}$ co-located in one child table of the task entity, and define each task label as the temporally-valid mean of the row-level product $a_j b_j$ over the linked child rows, computed strictly over rows dated before the task timestamp.  The same per-row cutoff column is consumed by all three models, so no model can access post-cutoff rows during labeling or inference. Column pairs touching value columns whose measurement unit varies across rows (\texttt{param\_value} and the confidence-interval limits, whose unit depends on a sibling code column) are excluded \emph{a priori} and outcome-blind, leaving nine child-table pairs.  These pairs form the x-axis labels of Figure \ref{fig:interaction_full} (\textit{right}).

We repeat the above experiment, this time explicitly adding an additional child-table column formed from this so-called ``golden feature'' interaction term.  This is meant to reflect a hypothetical scenario whereby an informed agent has supplied this product term as an additional useful feature.  We then explore the degree to which each RDB FM can effectively utilize this column to improve predictive performance in Figure \ref{fig:interaction_full} (\textit{right}).

\section{Details of Real-World Agentic Experiment} \label{app:agent_experiments}

Following the pointer at the end of Section \ref{sec:agents}, this experiment represents a preliminary proof-of-concept, not an exhaustive exploration of possibilities.  And certainly there are more principled ways of incorporating agents within a TFM-based paradigm than what we have done.  We also stress one additional point upfront:  Even as agents are often able to generate useful predictive features, this typically requires extensive domain knowledge and/or meta-data associated with a given RDB (this may also introduce new pathways for leakage and/or memorization that could artificially inflate performance; see Appendix \ref{app:LLM_usage}).  In contrast, TFM-based models like RDBLearn and TabPFN-Rel can in principle operate out of the box in new domains without such knowledge, providing greater flexibility and privacy.

\subsection{Experiment setup}
TFM-based methods leverage agent-friendly SQL-based RDB harnesses, where SQL queries are automatically generated to aggregate features from RDBs. These queries are agent-readable and provide an alternative entry point for understanding the predictive modeling task and core RDB structure. Based on this observation, we prompt Cursor Grok 4.5 under the default Cursor 3.6.31 harness. We provide a single prompt together with the raw training set and the corresponding features generated by the RDBLearn encoder $g_{\tiny \mbox{pf}}$. Note that we do not allow the agent to iterate (e.g., by using a loop). Therefore, the feature generation is a single forward pass. This is substantially different from the SOTA RelAgent \citep{huang2026relagent}, which trains numerous fully-supervised models (on a per-task basis) across multiple iterations of a multi-agent system.

For our experiments, the prompt for all tasks is the same except for the task name. As a representative example, the rel-trial prompt is as follows:

\begin{tcolorbox}[colback=white,colframe=gray!50!black,title=Prompt for Cursor Grok 4.5 to generate RDB features,]
If you look into what the rel-trial/study outcome tasks are, understand what the goal is, then look into DFS-depth 4 (including hop 1-4) SQLs. Can you analyze what features are useless and what new useful feature column you can add to the flattened table by SQL?"
\end{tcolorbox}

\subsection{Leakage Check}

For our experiment, only the training set is exposed to the agent. To investigate whether there is data leakage, we manually look into the trace and prompt the agent to self-check. As a result, two features for rel-event tasks were found aggregating future (the same week) data. They are \textit{sql\_hist\_ignore\_rate} and \textit{sql\_hist\_repeat\_rate}, which calculate past-week label statistics. Although the target predictive entity is excluded from the calculation, when this predictive entity correlates with a mid-week date, same-week future label information can still be included. (For reference, performance results in Table~\ref{tab:rdblearn-agent} below exclude these two features.) Beyond this, inspecting every tool call in the agent trace, we did not observe any behavior that would indicate retrieval/access of validation or testing set information.

\subsection{Results}
Table~\ref{tab:rdblearn-agent} presents the performance comparison between the agent-feature-enhanced RDBLearn (based on TabPFN-v3.5~\citep{jager2026tabpfn35}) and a SOTA multi-agent system, RelAgent~\citep{huang2026relagent}. We note that even without domain knowledge and metadata information, RDBLearn already outperforms RelAgent (Qwen3-32B). And when enhanced with agent-generated features, its performance is significantly boosted to be on-par with RelAgent (GPT-5.2).  Critically though, all RelAgent versions are \textit{fully-supervised}, training many per-task models, whereas RDBLearn remains a training-free FM enhanced with just a single batch of agent-generated features and no iterations.
\begin{table}[h]
\centering
\caption{RDBLearn+Agent results on RelBench-v1 binary classification tasks (AUROC$\uparrow$). Bold = best per column, underline = second best.  We emphasize that RelAgent is fully-supervised, training numerous models per task, while RDBLearn is an FM.}\label{tab:rdblearn-agent}
\resizebox{0.8\linewidth}{!}{%
\begin{tabular}{ll|cccc}
\toprule
\textbf{Dataset} & \textbf{Task} & \textbf{RelAgent (GPT-5.2)} & \textbf{RelAgent (Qwen3-32B)} & \textbf{RDBLearn} & \textbf{RDBLearn+Agent} \\
\midrule
\multirow{2}{*}{rel-amazon} & item-churn & \bm{$0.8284$} & $0.7992$ & $0.8270$ & \underline{$0.8273$} \\
                             & user-churn & \bm{$0.7078$} & $0.6922$ & \underline{$0.6973$} & $0.6968$ \\
\cmidrule(lr){1-6}
\multirow{2}{*}{rel-avito}  & user-clicks & $0.6836$ & $0.6510$ & \bm{$0.7050$} & \underline{$0.6994$} \\
                             & user-visits & \bm{$0.6779$} & $0.6462$ & \underline{$0.6714$} & $0.6661$ \\
\cmidrule(lr){1-6}
\multirow{2}{*}{rel-event}  & user-ignore & \bm{$0.8725$} & $0.7591$ & $0.7446$ & \underline{$0.7873$} \\
                             & user-repeat & \bm{$0.7820$} & $0.7473$ & $0.7646$ & \underline{$0.7753$} \\
\cmidrule(lr){1-6}
\multirow{2}{*}{rel-f1}     & driver-dnf  & \underline{$0.7834$} & $0.5767$ & $0.7263$ & \bm{$0.8351$} \\
                             & driver-top3 & \underline{$0.8523$} & $0.5804$ & $0.7975$ & \bm{$0.9204$} \\
\cmidrule(lr){1-6}
rel-hm                       & user-churn  & \bm{$0.7107$} & \underline{$0.7062$} & $0.6895$ & $0.6953$ \\
\cmidrule(lr){1-6}
\multirow{2}{*}{rel-stack}  & user-badge      & \bm{$0.8842$} & $0.6530$ & \underline{$0.8678$} & $0.8514$ \\
                             & user-engagement & \underline{$0.9041$} & $0.8911$ & \bm{$0.9051$} & $0.8982$ \\
\cmidrule(lr){1-6}
rel-trial                    & study-outcome & $0.7186$ & $0.7136$ & \underline{$0.7265$} & \bm{$0.7569$} \\
\midrule
\multicolumn{2}{l}{\textbf{AVG (12 tasks)}} & \underline{$0.7838$} & $0.7013$ & $0.7602$ & \bm{$0.7841$} \\
\bottomrule
\end{tabular}
}
\end{table}

\section{Technical Proofs} \label{app:proofs}

\subsection{Proof of Proposition \ref{prop:locality_stuff}}

Generalizing from \citet{nagler2023statistical}, the proof here is relatively straightforward once the RDB-dependent setup is specified through Definition \ref{def:reverse_time}, the assumption that (\ref{eq:ideal_asymptotic_conditions}) holds, and that  $p\big(y_i | \calG_H(\bx_i) \big) = p\big(y_i | \bz_i \big)$ $\forall i=0,\ldots,n$.  The first and third assumption alone imply that a given label $y_i$ is independent of others when conditioned on the corresponding encoder outputs, i.e., $p(y_i, y_j | \bz_i, \bz_j) = p(y_i|\bz_i)p(y_j|\bz_j) ~\forall i,j$.  Hence we may also infer that for any labeled sample, the conditional $p(y | \bz)$, as opposed to merely $p\big(y| \calG_H(\bx) \big)$, is stable across $n$, even while encoder representations $\bz$ need not be.

Now consider any sequence pairs $p,\check{p} \in \calP$, along with their associated label generators $p(y|\bz)$ and $\check{p}(y|\bz)$ respectively, satisfying $p(y| \bz) = \check{p}(y|\bz)$  whenever $\|\bz_{\tiny \mbox{test}} - \bz \| < \epsilon$ for any fixed $\epsilon > 0$.  For all such pairs, we have from (\ref{eq:ideal_asymptotic_conditions}) (bias part) that
\begin{eqnarray} \label{eq:mean_convergence_cond}
&& \hspace*{-0.8cm} p(y_{\tiny \mbox{test}} | \bz_{\tiny \mbox{test}}) - \mathbb{E}_{\calD_n \sim p_n }\left[ q_\theta\big(y_{\tiny \mbox{test}} ~|~\{\bz_{i}, y_i \}_{i=1}^n, \bz_{\tiny \mbox{test}} \big) \right]   \\
&& \hspace*{2cm} - ~~ \Big( \check{p}(y_{\tiny \mbox{test}} | \bz_{\tiny \mbox{test}}) - \mathbb{E}_{\calD_n \sim \check{p}_n }\left[ q_\theta\big(y_{\tiny \mbox{test}} ~|~\{\bz_{i}, y_i \}_{i=1}^n, \bz_{\tiny \mbox{test}} \big) \right]  \Big) \nonumber \\
&&  \mathbb{E}_{\check{\calD}_n \sim \check{p}_n }\left[ q_\theta\big(y_{\tiny \mbox{test}} ~|~\{\check{\bz}_{i}, \check{y}_i \}_{i=1}^n, \bz_{\tiny \mbox{test}} \big) \right] - \mathbb{E}_{\calD_n \sim p_n }\left[ q_\theta\big(y_{\tiny \mbox{test}} ~|~\{\bz_{i}, y_i \}_{i=1}^n, \bz_{\tiny \mbox{test}} \big) \right] ~~ \rightarrow ~~ 0 \nonumber 
\end{eqnarray}
as $n \rightarrow \infty$, where $\{\bz_{\tiny \mbox{test}}, y_{\tiny \mbox{test}}\}$ is treated as a fixed evaluation point.  Combining with (\ref{eq:ideal_asymptotic_conditions}) (variance part) we then may conclude that
\begin{equation}
  \left|  q_\theta\big(y_{\tiny \mbox{test}} ~|~\{\check{\bz}_{i}, \check{y}_i \}_{i=1}^n, \bz_{\tiny \mbox{test}} \big) - q_\theta\big(y_{\tiny \mbox{test}} ~|~\{\bz_{i}, y_i \}_{i=1}^n, \bz_{\tiny \mbox{test}} \big) \right| ~~\rightarrow ~~ 0.
\end{equation}
From standard diagonalization arguments, the same holds true for some $\epsilon_n \rightarrow 0$ since no assumptions were placed on the original $\epsilon$, therefore (\ref{eq:local_error}) is established.   \myendofproof

\subsection{Proof of Proposition \ref{prop:interpolation_limitation}}

We show this result through an interpretable counter-example.  Note that there is nothing particularly special about the configuration we introduce for this purpose; rather it merely represents a simplified case sufficient for illustration purposes and completing the proof.  As we will see, the core ingredient is the existence of at least some degree of nonlinear interaction between relational effects within $\calG_H(\bx)$ and the original tabular features $\bx$.

\paragraph{Basic Setup.} For any $\varepsilon \in (0,1)$ and $C > 0$ we define mappings $f'_1: \mathbb{R}^{d} \rightarrow \mathbb{R}^2$, $f'_2 : \mathbb{G}_H \rightarrow \mathbb{R}^2$, and $\Psi' : \mathbb{R}^2 \rightarrow \mathbb{R}$ as
\begin{equation} \label{eq:gen_process_eqs}
f'_1: ~ \left[\begin{array}{c} 
    u_1 \\ 0
    \end{array} \right] = \left[\begin{array}{c} 
    \bomega^\top \widetilde{\bx}^{\tiny \mbox{real}} \\ 0
    \end{array} \right] +  \left[\begin{array}{c} 
    \beta \\ 0
    \end{array} \right],  \hspace*{0.5cm} f'_2: ~ \left[\begin{array}{c} 
    0 \\ u_2
    \end{array} \right] = \left[\begin{array}{c} 
    0 \\ \pi\left(\calG_H(\bx) \backslash \bx \right)
    \end{array} \right],
\end{equation}
\begin{equation}
\Psi': ~ y = w v_1 \mbox{sign}[v_2] + v_1 + v_2, \nonumber
\end{equation}
where $\bomega \in \mathbb{R}^{d'}$, $\beta \in \mathbb{R}$, and $w \in \mathbb{R}$ are constants in general position, and $\pi : \mathbb{G}_H \rightarrow \mathbb{R}$.  For the latter, we choose $\pi$ such that under $p_\tau$ we have
\begin{equation}
    \mbox{Prob}\left[\pi\big( \calG_H(\bx_{\tiny \mbox{test}})\backslash \bx_{\tiny \mbox{test}} \big) > 0 \right] ~~=~~ \mbox{Prob}\left[\pi\big( \calG_H(\bx_{\tiny \mbox{test}})\backslash \bx_{\tiny \mbox{test}} \big) < 0 \right] ~~ = ~~ 0.5.
\end{equation}
Additionally, underpinning (\ref{eq:gen_process_eqs}) we use $\mathbb{G}_H$ to denote the relational neighborhood space of $\calG_H(\bx) \backslash \bx$, while $\{v_1,v_2\}$ and $\{u_1,u_2\}$ are the variables adopted to express the relevant input and output spaces, respectively.  Combining these specifications with (\ref{eq:label_function2}) we arrive at the label generation function
\begin{equation}
    y = w (1-\tau)\tau  u_1 \mbox{sign}[u_2] + (1-\tau) u_1 + \tau u_2,
\end{equation}
where we observe nonlinear coupling between features derived from $\bx$ (expressed through $u_1$) and $\calG_H(\bx) \backslash \bx$ (expressed through $u_2$).

\paragraph{Handling  Equalities in (\ref{eq:prop_interpolation_conditions}).}  To establish the first equality of (\ref{eq:prop_interpolation_conditions}), we observe that by construction $y = \bomega^\top \widetilde{\bx}^{\tiny \mbox{real}} + \beta$ when $\tau = 0$.  However, we may infer that
\begin{equation} \label{eq:equality_w_zero}
    \min_{f_\theta^{\tiny \mbox{tab}} \in \calF} \mathbb{E}_{\calD_0 \sim p_0} \Big[~ \big| y_{\tiny \mbox{test}} -  \hat{y}_{\tiny \mbox{test}}^{\tiny \mbox{tab}}  \big|~ \Big] ~ = ~ \min_{f_\theta^{\tiny \mbox{tab}} \in \calF } \mathbb{E}_{\calD_0 \sim p_0} \Big[~ \big| \bomega^\top \widetilde{\bx}^{\tiny \mbox{real}}_{\tiny \mbox{test}} + \beta -  f_\theta^{\tiny \mbox{tab}}\big(\{\bx_{i}, y_i \}_{i=1}^n, \bx_{\tiny \mbox{test}} \big)  \big|~ \Big] ~=~ 0,
\end{equation}
where $p_0$ follows Definition \ref{def:RDB_gen_process} with $\tau = 0$.  Given the assumption that $n>d'$, the equality with zero in (\ref{eq:equality_w_zero}) occurs by straightforward extension of  \citet{JUICE_paper_2026}[Proposition 4.2], which is not operationally effected by padding $\bx_{\tiny \mbox{test}}$ and each $\bx_i$ with additional categorical elements.  Meanwhile, the second equality in (\ref{eq:prop_interpolation_conditions})  stems from the fact that $y = \pi\big( \calG_H(\bx )\backslash \bx \big)$ and our encoder constraint set $\bar{g}$ allows for modeling an arbitrary function $\pi$ over the relational neighborhood excluding $\bx$.  Hence when $\tau = 1$ we trivially have that
\begin{eqnarray} \label{eq:equality_w_zero2}
    && \hspace*{-0.5cm} \min_{f_\theta^{\tiny \mbox{rdb}} \in \calF,~ g_\phi \in \bar{g}} \mathbb{E}_{\calD_1 \sim p_1} \Big[~ \big| y_{\tiny \mbox{test}} -  \hat{y}_{\tiny \mbox{test}}^{\tiny \mbox{rdb}}  \big|~ \Big]  \\
    && \hspace*{1.0cm}  \equiv ~~ \min_{f_\theta^{\tiny \mbox{rdb}} \in \calF,~ g_\phi \in \bar{g}} \mathbb{E}_{\calD_1 \sim p_1} \Big[~ \big| \pi\big( \calG_H(\bx_{\tiny \mbox{test}})\backslash \bx_{\tiny \mbox{test}} \big) -  f_\theta^{\tiny \mbox{rdb}}\big(\{\bz_{i}, y_i \}_{i=1}^n, \bz_{\tiny \mbox{test}} \big)  \big|~ \Big] ~=~ 0. \nonumber
\end{eqnarray}

\paragraph{Handling the Inequality in (\ref{eq:prop_interpolation_conditions}).}  Lastly, we handle the third line of (\ref{eq:prop_interpolation_conditions}), namely, the inequality that results for $\tau$ values bounded away from zero and one.  We establish this bound via several steps as follows.  We also remark upfront that $\alpha$ can be absorbed into the specifications of $\hat{y}_{\tiny \mbox{test}}^{\tiny \mbox{tab}}$ and $\hat{y}_{\tiny \mbox{test}}^{\tiny \mbox{tab}}$ without loss of generality.

First, because of the constraint $g_\phi \in \bar{g}$, the encoder cannot reliable reconstruct $\bx^{\tiny \mbox{real}}_{\tiny \mbox{test}}$; at most we can guarantee reconstruction of $d-1$ coordinates per Definition \ref{def:compressed_embedding}.  Moreover, it is not possible to reconstruct the missing coordinate from the others with a continuous mapping from $\calF$, as doing so would violate the assumption from Definition \ref{def:RDB_gen_process} that any $p(\bx^{\tiny \mbox{real}})$ is bounded from above (i.e., if we could reconstruct this coordinate with a continuous mapping, it would imply that the data lies on a low-dimensional manifold within $\mathbb{R}^d$, which by necessity would have infinite density).

Given these considerations, we can again invoke \citet{JUICE_paper_2026}[Proposition 4.2] to infer that there will be non-negligible error (in expectation over the generation process) for any attempt to predict $u_{1,{\tiny \mbox{test}}}$ from available context $\{\bz_{i}, y_i \}_{i=1}^n$ and $\bz_{\tiny \mbox{test}}$.  Intuitively, this is because the support pattern for extracting $\widetilde{\bx}^{\tiny \mbox{real}}_{\tiny \mbox{test}}$ from $\bx^{\tiny \mbox{real}}_{\tiny \mbox{test}}$ varies uniformly across random draws from $p_\tau$, and so in expectation there will always be cases where needed information is missing.  From this we may conclude that
\begin{equation}
    \inf_{f_\theta^{\tiny \mbox{rdb}} \in \calF,~g_\phi \in \bar{g}} \mathbb{E}_{\calD_\tau \sim p_\tau} \big[~ \big| u_{1,{\tiny \mbox{test}}} -  \hat{y}_{\tiny \mbox{test}}^{\tiny \mbox{rdb}}  \big| \big] ~\geq~ \delta
    \end{equation}
for any $\tau \in [\varepsilon, 1 - \varepsilon]$ and some $\delta > 0$.  But this will also remain true when conditioning on the corresponding $u_{2,{\tiny \mbox{test}}}$.  This is because $u_{2,{\tiny \mbox{test}}}$ is merely a function of other rows within $\bX$, and conditioning on any or all rows (or functions thereof), excluding the current $\bx_{\tiny \mbox{test}}$, will not remove the prediction error, since these rows are generated IID by Definition \ref{def:RDB_gen_process}.  And so we have that
\begin{equation}
    \inf_{f_\theta^{\tiny \mbox{rdb}} \in \calF,~g_\phi \in \bar{g}} \mathbb{E}_{\calD_\tau \sim p_\tau} \big[~ \big| u_{1,{\tiny \mbox{test}}} -  \hat{y}_{\tiny \mbox{test}}^{\tiny \mbox{rdb}}  \big|~~\big| ~ u_{2,{\tiny \mbox{test}}} \geq \lambda ~ \big] ~\geq~ \delta'
\end{equation}
for some possibly new $\delta' > 0$; an analogous error term is bounded away from zero when conditioning on the event $u_{2,{\tiny \mbox{test}}} < 0$.  And therefore from the tower rule of conditional expectations, we also have
    \begin{equation}
    \inf_{f_\theta^{\tiny \mbox{rdb}},~g_\phi \in \bar{g}} \mathbb{E}_{\calD_\tau \sim p_\tau} \big[~ \big| u_{1,{\tiny \mbox{test}}} \mbox{sign}\big[u_{2,{\tiny \mbox{test}}}\big]  -  \hat{y}_{\tiny \mbox{test}}^{\tiny \mbox{rdb}}  \big| ~ \big] ~\geq~ \delta'' ~>~ 0,
    \end{equation}
a lower bound that becomes $w(1-\tau)\tau \delta''$ once we reintroduce scale factors.  If we choose $w = \frac{C}{(1-\varepsilon)\varepsilon \delta''}$, then
\begin{equation}
w(1-\tau)\tau \delta'' ~~ = ~~ \frac{C(1-\tau)\tau }{(1-\varepsilon)\varepsilon} ~~ \geq ~~ C  ~~~~~\forall \tau \in [\varepsilon, (1-\varepsilon)].
\end{equation}
Finally, the remaining factors $(1-\tau) u_{1,{\tiny \mbox{test}}}$, $\tau u_{2,{\tiny \mbox{test}}}$, and $\hat{y}_{\tiny \mbox{test}}^{\tiny \mbox{tab}}$ from within the last expectation of (\ref{eq:prop_interpolation_conditions}) cannot appreciably improve the lower bound of $C$.  Note that the factor $\tau u_{2,{\tiny \mbox{test}}}$ can be absorbed into $\hat{y}_{\tiny \mbox{test}}^{\tiny \mbox{rdb}}$ without loss of generality; they share the same input range while the former is a pre-specified function and the latter is arbitrary.  

Analogously, w.l.o.g.~$(1-\tau) u_{1,{\tiny \mbox{test}}}$ can be absorbed into $\hat{y}_{\tiny \mbox{test}}^{\tiny \mbox{tab}}$.  But the flexibility from $\hat{y}_{\tiny \mbox{test}}^{\tiny \mbox{tab}}$ itself is inconsequential to altering the bound. This is because $u_{1,{\tiny \mbox{test}}} \indep u_{2,{\tiny \mbox{test}}}$, given that $\bx_{\tiny \mbox{test}}^{\tiny \mbox{real}} \indep \calG_H(\bx)\backslash \bx$ per Definition \ref{def:RDB_gen_process}, and so reducing the error approximating the interaction term $u_{1,{\tiny \mbox{test}}} \mbox{sign}[u_{2,{\tiny \mbox{test}}}]$ is not possible without access to $u_{2,{\tiny \mbox{test}}}$ (noting that $u_{2,{\tiny \mbox{test}}}$ has a median of zero).  And $\hat{y}_{\tiny \mbox{test}}^{\tiny \mbox{tab}}$ cannot generally reproduce $u_{2,{\tiny \mbox{test}}}$, since the latter will depend (in expectation at least) on IID context rows outside of the ICL set mitigated through choice of the function $\pi$.  \myendofproof

\subsection{Proof of Proposition \ref{prop:labels_as_features}} \label{app:proof1}

\paragraph{High-Level Intuition.}  To illustrate the gist of this result, assume that the ground-truth label $y_{\tiny \mbox{test}}$ for any $\bx_{\tiny \mbox{test}}$ is given by a parity check on available labels in $\calG_H^*(\bx_{\tiny \mbox{test}})$.  This labeling rule is impossible to determine from $y_H(\bx_{\tiny \mbox{test}})$ alone, since for any node $\bx'$ associated with a label $y' \in y_H(\bx_{\tiny \mbox{test}})$, the corresponding set $y_H(\bx')$ need not be fully observed, i.e., the overlap between $y_H(\bx')$ (with a known label for $\bx'$) and $y_H(\bx_{\tiny \mbox{test}})$ (with an unknown label for $\bx_{\tiny \mbox{test}}$) is only partial.  Hence the parity rule is not identifiable, as even a single missing label can flip the result.  This disrupts inductive inference from the information available within $\calG^*_H(\bx_{\tiny \mbox{test}})$, which is all a fixed FM encoder has access too.  In contrast, if per-dataset training is allowed over RDB instances formed via a given parity-based labeling rule, it is straightforward to learn the associated mapping on a dataset-by-dataset or task-by-task basis.  We remark that, although the proof below relies on parity checks for illustrative convenience, the underlying phenomena is emblematic of quite broad scenarios where diverse heterophilic (as opposed to homophilic) relationships are dominant.

\paragraph{Basic Setup.}  The basic proof follows from the construction of a suitable joint distribution over context  $\widetilde{\calG}_H^*(\bx_{\tiny \mbox{test}})$ and test label $y_{\tiny \mbox{test}}$.  Meanwhile, by assumption we have that  $\calG_H^*(\bx_{\tiny \mbox{test}})$ is fixed and given.  To begin, we introduce a latent indicator variable $z \in \{0,1\}$ that selects among two deterministic generative processes for $\{\widetilde{\calG}^*_H(\bx_{\tiny \mbox{test}}), y_{\tiny \mbox{test}} \}$.  In this way, all randomness in the joint distribution $p(\widetilde{\calG}^*_H(\bx_{\tiny \mbox{test}}), y_{\tiny \mbox{test}} ~|~\calG_H^*(\bx_{\tiny \mbox{test}}))$ will ultimately reduce to randomness from $p(z)$.  We emphasize that this is not meant to exclude broader sources of randomness that naturally describe RDB content; instead, we are merely distilling down to a minimal design case that nonetheless still allows us to establish the proposition.  Other latent sources of variation associated with real-world RDBs then become largely orthogonal to our result and need not be explicitly specified.

\paragraph{Label Assumptions.}  We assume that all labels are binary, namely, $y \in \{0,1\}$.  While generalization to broader label domains $\calY$ is feasible, it introduces superfluous complexity given that the binary case is already sufficient for establishing the proof.  For any $y_H(\bx) \neq \emptyset$, where $y_H(\bx)$ indicates the set of all past labels present within $\calG^*_H(\bx)$ for some arbitrary $\bx$, we define conditional label generation via a parity function and its negative given by
\begin{eqnarray} \label{eq:label_assign}
\pi\big[ y_H(\bx) ~;~z = 1 \big] & := & \left[ \sum_{y \in y_H(\bx)} y \right] \hspace*{-0.3cm} \mod 2 \nonumber \\
\pi\big[ y_H(\bx) ~;~z = 0 \big] & := & \neg \pi \big[ y_H(\bx); z = 1 \big],
\end{eqnarray}
both of which output a label $y \in \{0,1\}$.

\paragraph{Context Assumptions.}  We specify a function $h\big( \calG_H^*(\bx_{\tiny \mbox{test}}) ; z\big)$ for context generation.  To ensure no conflicting labels, for $z=1$ we define this $h$ such that $\widetilde{\calG}^*_H(\bx_{\tiny \mbox{test}})$ has label assignments consistent with both $y_H(\bx_{\tiny \mbox{test}})$ and the label assignment rule $\pi\big[ ~ \cdot ~;~z = 1 \big]$ from (\ref{eq:label_assign}).  This will always be achievable by introducing new appropriately-labeled neighbors to specific nodes of $\calG_H^*(\bx_{\tiny \mbox{test}})$ associated with elements of $y_H(\bx_{\tiny \mbox{test}})$.   

To establish this possibility, assume an arbitrary trial arrangement of node features, edges, and labels within $\widetilde{\calG}^*_H(\bx_{\tiny \mbox{test}})$, but following the typing of $\calG^*_H(\bx_{\tiny \mbox{test}})$ for consistency. Then consider any $y' \in y_H(\bx_{\tiny \mbox{test}})$ with corresponding $\bx'$ extracted from the associated row of $\bX \equiv \bT^K$ (per original definitions in the main text).  For consistency with the labeling rule (\ref{eq:label_assign}) with $z=1$, we require that $y' = \pi\big[ y_H(\bx') ~;~z = 1 \big]$, which may not initially be the case since labeling assignments were thus far assumed arbitrary.  However, for any node where this is not the case, we can add a single additional labeled neighbor to $\widetilde{\calG}^*_H(\bx_{\tiny \mbox{test}})$, with an earlier time-stamp and a label selected to flip the parity check, such that we obtain the necessary equality.  

Such a neighbor can be added in multiple ways given that $H \geq 2$ (note that if $H=1$ then $|y_H(\bx_{\tiny \mbox{test}})| = 0$, i.e., no labels within one hop, which is not allowed by assumption).  First, assume $H=2$, if some table $T^k$ exists with at least two FK columns pointing to the PK column associated with $T^K \equiv \bX$, we can trivially add such a neighbor by adding a new labeled row to $T^K$ and a new row to $T^k$ that links the former to $\bx' \in T^K$.  Meanwhile, if such a $T^k$ does not exist, we can simply add such a table to $\widetilde{\calG}^*_H(\bx_{\tiny \mbox{test}})$.  Moreover, when supplied with an earlier time-stamp, the added labeled node/row need not introduce any additional constraints vis-\`a-vis  (\ref{eq:label_assign}). With regard to the latter, if $\bx''$ indicates the added node, we can directly enforce $y_H(\bx'') = \emptyset$ as no nodes with earlier time-steps exist, and so $y''$ can be selected free of any constraint from (\ref{eq:label_assign}), which would only apply if $y_H(\bx'') \neq \emptyset$.  Lastly, because this added label will by design be two hops from $\bx'$ and three or more hops from any other labeled rows, it will not introduce new conflicts per (\ref{eq:label_assign}) if $H =2$.  For $H > 2$ an analogous procedure can be applied pushing the new labeled node to $H$ hops away from $\bx'$ so as to achieve the same effect.

In an analogous manner we can also define $h$ when $z=0$, handling label consistency via the introduction of neighboring nodes to adjust parity accordingly.  Again, we are granted this flexibility by assumption, since only $\calG^*_H(\bx_{\tiny \mbox{test}})$ is fixed and given.

\paragraph{Finalizing Part 1.}  Per the assumptions stated above and the appropriate counting measure, it directly follows that $\int p\left(y_{\tiny \mbox{test}} | \calG^*_{\tiny \mbox{test}} \right) \log p\left(y_{\tiny \mbox{test}} | \calG^*_{\tiny \mbox{test}} \right) d y_{\tiny \mbox{test}} ~=~0$ trivially establishing that $y_{\tiny \mbox{test}}$ is a deterministic function of $\calG^*_{\tiny \mbox{test}}$.  Furthermore, given $\calG^*_{\tiny \mbox{test}} = \left\{\widetilde{\calG}^*_H(\bx_{\tiny \mbox{test}}), \calG^*_H(\bx_{\tiny \mbox{test}}) \right\}$ by definition, we also have that 
\begin{eqnarray} \label{eq:KL_stuff1}
&& \hspace*{-0.6cm} \mathbb{E}_{p\left(\widetilde{\calG}^*_H(\bx_{\tiny \mbox{test}}) | \calG^*_H(\bx_{\tiny \mbox{test}}) \right)} \left[ \mathbb{KL}\Big[ p(y_{\tiny \mbox{test}} |\calG^*_{\tiny \mbox{test}})  ~||~ q_\theta\big(y_{\tiny \mbox{test}} ~|~ g_\phi\big[ \calG^*_H(\bx_{\tiny \mbox{test}}) \big]  \big) \Big] \right] \nonumber  \\
&& \hspace*{-0.5cm} = \int p\left(\widetilde{\calG}^*_H(\bx_{\tiny \mbox{test}}) | \calG^*_H(\bx_{\tiny \mbox{test}}) \right)
\int p\left(y_{\tiny \mbox{test}} | \calG^*_{\tiny \mbox{test}} \right) \log \frac{p\left(y_{\tiny \mbox{test}} | \calG^*_{\tiny \mbox{test}} \right)}{q_\theta\big(y_{\tiny \mbox{test}} ~|~ g_\phi\big[ \calG^*_H(\bx_{\tiny \mbox{test}}) \big]  \big)} d y_{\tiny \mbox{test}} d \widetilde{\calG}^*_H(\bx_{\tiny \mbox{test}})   \nonumber \\
&& \hspace*{-0.5cm} = - \int p\left(\widetilde{\calG}^*_H(\bx_{\tiny \mbox{test}}) | \calG^*_H(\bx_{\tiny \mbox{test}}) \right)
\int p\left(y_{\tiny \mbox{test}} | \widetilde{\calG}^*_H(\bx_{\tiny \mbox{test}}), \calG^*_H(\bx_{\tiny \mbox{test}})   \right) \log q_\theta\big(y_{\tiny \mbox{test}} ~|~ g_\phi\big[ \calG^*_H(\bx_{\tiny \mbox{test}}) \big]  \big) d y_{\tiny \mbox{test}} d \widetilde{\calG}^*_H(\bx_{\tiny \mbox{test}})   \nonumber \\
&& \hspace*{-0.5cm} = - \int \int p\left(\widetilde{\calG}^*_H(\bx_{\tiny \mbox{test}}),y_{\tiny \mbox{test}} | \calG^*_H(\bx_{\tiny \mbox{test}}) \right) \log q_\theta\big(y_{\tiny \mbox{test}} ~|~ g_\phi\big[ \calG^*_H(\bx_{\tiny \mbox{test}}) \big]  \big) d y_{\tiny \mbox{test}} d \widetilde{\calG}^*_H(\bx_{\tiny \mbox{test}})  
\end{eqnarray}

for any fixed $g_\phi$ and $q_\theta$.   By construction, $p\left(\widetilde{\calG}^*_H(\bx_{\tiny \mbox{test}}),y_{\tiny \mbox{test}} | \calG^*_H(\bx_{\tiny \mbox{test}}) \right) \neq 0$ at only two points associated with $z = 0$ and $z = 1$.  Let $y_{\tiny \mbox{test}}^{(0)}$ and $y_{\tiny \mbox{test}}^{(1)}$ denote the value of $y_{\tiny \mbox{test}}$ at these two points.  Moreover, if we assume that $z$ is drawn uniformly over $\{0,1\}$, we have $\calG_H^*(\bx_{\tiny \mbox{test}}) \indep z$ and so we may conclude from (\ref{eq:KL_stuff1}) that
\begin{eqnarray}
&& \hspace*{-1.5cm} \mathbb{E}_{p\left(\widetilde{\calG}^*_H(\bx_{\tiny \mbox{test}}) | \calG^*_H(\bx_{\tiny \mbox{test}}) \right)} \left[ \mathbb{KL}\Big[ p(y_{\tiny \mbox{test}} |\calG_{\tiny \mbox{test}})  ~||~ q_\theta\big(y_{\tiny \mbox{test}} ~|~ g_\phi\big[ \calG^*_H(\bx_{\tiny \mbox{test}}) \big]  \big) \Big] \right] \nonumber \\
&& \hspace*{-0.5cm} =  ~~ -p(z=0)\log q_\theta\big(y_{\tiny \mbox{test}}^{(0)} ~|~ g_\phi\big[ \calG^*_H(\bx_{\tiny \mbox{test}}) \big]  \big) ~-~ p(z=1)\log q_\theta\big(y_{\tiny \mbox{test}}^{(1)} ~|~ g_\phi\big[ \calG^*_H(\bx_{\tiny \mbox{test}}) \big]  \big) \nonumber \\
&& \hspace*{-0.5cm} \geq ~~  \mathbb{H}[z] ~ = ~ \log 2.
\end{eqnarray}
This lower bound is achievable iff $q_\theta\big(y_{\tiny \mbox{test}} ~|~ g_\phi\big[ \calG^*_H(\bx_{\tiny \mbox{test}}) \big]  \big)$ is a uniform distribution over $\{0,1\}$.

\paragraph{Finalizing Part 2.}  Assume that some parameter-free encoder $g_{\tiny \mbox{pf}}$ computes the element-wise sum of $y_H(\bx)$; other arbitrary features may be computed as well, but they are are ultimately treated as spurious features in what follows.  Given this setup, the prediction head $q_\theta$ need only learn to differentiate even and odd sums associated with $y_H(\bx)$, while ignoring all other feature columns.  For example, if $z=1$ governs the generative process, then odd values of any sum over $y_H(\bx)$ will be mapped to $1$, even values to 0.  Such functions can be directly represented by ReLU networks, and while not a requirement for this proof, are even efficiently learnable with additional assumptions \citep{daniely2020learning}.  \myendofproof

\subsection{Proof of Proposition \ref{prop:labels_and_feature_importance}}

\paragraph{High-Level Intuition.}  The proof follows through modifications applied to the original parity setup from Appendix \ref{app:proof1}.   For illustrative simplicity, assume a single informative binary RDB feature column, whereby ground-truth labels are a deterministic function, akin to a parity check, on the values of all features from this column contained within $\calG^*_H(\bx_{\tiny \mbox{test}})$.  All other RDB columns are also binary but uninformative.  Inclusion of $y_H(\bx_{\tiny \mbox{test}})$ within $\calG^*_H(\bx_{\tiny \mbox{test}})$ does not allow for differentiating which column is informative, for analogous reasons to those presented in establishing Proposition \ref{prop:labels_as_features}.  In brief,
for any node $\bx'$ associated with a label $y' \in y_H(\bx_{\tiny \mbox{test}})$, the corresponding set $y_H(\bx')$ \textit{and its associated features} need not be fully observed, and so a parity function on \textit{any} column can produce an arbitrary distribution of $y_H(\bx_{\tiny \mbox{test}})$.   Consequently, the parity rule is not identifiable regardless of whether $y_H(\bx_{\tiny \mbox{test}})$ is present or not as an encoder input.  Note that this line of reasoning is not restricted to binary feature columns, as we can quantize continuous or categorical columns into binary ones in forming deterministic labeling functions for analysis purposes.  We now turn to the actual proof.

\paragraph{Basic Setup.}    As in the proof of Proposition \ref{prop:labels_as_features}, all randomness is encapsulated by a binary latent variable $z$ that selects between two deterministic generative processes, now over $\{\widetilde{\calG}^*_H(\bx_{\tiny \mbox{test}}), y_H(\bx_{\tiny \mbox{test}}), y_{\tiny \mbox{test}} \}$, i.e., only $\calG_H(\bx_{\tiny \mbox{test}})$ as opposed to $\calG^*_H(\bx_{\tiny \mbox{test}})$ is assumed fixed/given.  When granted $\calG_H(\bx)$ for some arbitrary $\bx$, we specify the conditional label generation functions
\begin{eqnarray} \label{eq:label_assign2}
\pi\big[ \calG_H(\bx) ~;~z = 1 \big] & := & \left[ \sum_{\{i,j,k\} \in S_H^{(1)}(\bx)} \beta\left( t_{ij}^k \right) \right] \hspace*{-0.3cm} \mod 2 \nonumber \\
\pi\big[ \calG_H(\bx) ~;~z = 0 \big] & := & \left[ \sum_{\{i,j,k\} \in S_H^{(0)}(\bx)} \beta\left( t_{ij}^k \right) \right] \hspace*{-0.3cm} \mod 2,
\end{eqnarray}
where $S_H^{(1)}(\bx)$ and $S_H^{(0)}(\bx)$, associated with $z=1$ and $z=0$ respectively, represent distinct support sets over RDB column elements reachable within $H$ hops from $\bx$ following PK-FK pairs.  In this way any $\{i,j,k\} \in S^{(1)}_H(\bx)$ (or analogously, within $S^{(0)}_H(\bx)$) indicates the $i$-th element within a specific RDB column $\bt_{:j}^k$, namely column $j$ in table $k$.  Meanwhile $\beta$ denotes a mapping from each table element to a binary value (this discretization is by no means necessary, but it facilitates a simplified exposition that is sufficient for establishing the core result).  These values are then subsequently combined via the stated parity functions in (\ref{eq:label_assign2}) to produce binary labels $y \in \{0,1\}$.

\paragraph{Establishing Part 1.}  By adopting (\ref{eq:label_assign2}) as the label generating function, we trivially achieve the conditions of part 1 in Proposition \ref{prop:labels_and_feature_importance}.  Specifically, with all randomness deferred to $z$, which explicitly fixes the support set, all labels default to a deterministic function of the chosen support set, either $S_H^{(0)}(\bx)$ or $S_H^{(1)}(\bx)$ for any $\bx$.

\paragraph{Establishing Part 2.}  If we stipulate that $z$ is drawn uniformly over $\{0,1\}$ independently of $ \calG_H(\bx_{\tiny \mbox{test}})$, then we have
\begin{equation}
    p\left(S = S_H^{(0)}(\bx_{\tiny \mbox{test}}) ~|~  \calG_H(\bx_{\tiny \mbox{test}}) \right) ~=~ p\left(S = S_H^{(1)}(\bx_{\tiny \mbox{test}}) ~|~  \calG_H(\bx_{\tiny \mbox{test}}) \right) ~=~ \tfrac{1}{2}.
\end{equation}
However, by design we also have the flexibility to select the distribution over $\widetilde{\calG}^*_H(\bx_{\tiny \mbox{test}})$ such that $y_H(\bx_{\tiny \mbox{test}})$ is invariant to $z$, even while $y_{\tiny \mbox{test}}$ still depends on $z$ through (\ref{eq:label_assign2}) applied at $\bx_{\tiny \mbox{test}}$.  To see this, note that each $y' \in y_H(\bx_{\tiny \mbox{test}})$ is computed as $\pi[\calG_H(\bx'); z]$, where $\bx'$ is the target table feature row associated with $y'$.  Hence we only need enforce 
\begin{equation} \label{eq:needed_equality}
    \pi\big[\calG_H(\bx');~ z=1 \big] = \pi\big[\calG_H(\bx');~ z=0 \big] ~~~ \mbox{for each} ~~ y' \in y_H(\bx_{\tiny \mbox{test}}).
\end{equation}
But this will always be possible per the following:  Suppose the equality from (\ref{eq:needed_equality}) does not initially hold.  We may add an additional element  to either $S_H^{(1)}(\bx')$ or $S_H^{(0)}(\bx')$ (but not both) such that the corresponding tuple produces $\beta(t_{ij}^k) = 1$ to flip the parity and induce equality.  This is achievable without disrupting equality elsewhere by placing the added element within $\widetilde{\calG}^*_H(\bx_{\tiny \mbox{test}})$, at a location $H$ hops from $\bx'$ and greater than $H$ hops from other elements of $y_H(\bx_{\tiny \mbox{test}})$.  This placement mirrors an analogous placement strategy applied in Appendix \ref{app:proof1}, only here labeled nodes are not required.

Lastly, if (\ref{eq:needed_equality}) holds, then
\begin{equation}
   p\big(S  | \calG^*_H(\bx_{\tiny \mbox{test}}) \big) = p\big(S  | \calG_H(\bx_{\tiny \mbox{test}}), y_H(\bx_{\tiny \mbox{test}}) \big) = p\big(S | \calG_H(\bx_{\tiny \mbox{test}}) \big),
\end{equation}
completing the demonstration of part 2.   \myendofproof

\end{document}